\documentclass[lettersize,journal]{IEEEtran}
\usepackage{amsmath,amsfonts}
\usepackage{algorithmic}
\usepackage{algorithm}
\usepackage{array}
\usepackage[caption=false,font=normalsize,labelfont=sf,textfont=sf]{subfig}
\usepackage{textcomp}
\usepackage{stfloats}
\usepackage{url}\usepackage{bbding}
\usepackage{verbatim}
\usepackage{graphicx}
\usepackage{adjustbox}
\usepackage{booktabs,multirow}
\usepackage{colortbl}
\usepackage{makecell,rotating}
\usepackage{bm}
\usepackage{verbatim}
\usepackage{color}
\usepackage{arydshln}
\usepackage{makecell,rotating}
\usepackage{array}
\usepackage{amsmath}
\usepackage{boldline}
\usepackage{booktabs} 
\usepackage{amssymb}
\usepackage{amsthm}
\usepackage{graphicx}
\usepackage{cite}
\usepackage{color}
\usepackage{epstopdf}
\usepackage{subeqnarray}
\usepackage{cases}
\usepackage{caption}
\usepackage{cite}
\usepackage{graphicx}
\usepackage{wrapfig}
\usepackage{colortbl}
\usepackage{float}
\usepackage{overpic}
\usepackage{multirow}
\usepackage{booktabs}
\usepackage[dvipsnames]{xcolor}
\definecolor{citecolor}{RGB}{119,185,0}
\definecolor{citecolor1}{RGB}{66,168,235}
\definecolor{BrickRed}{RGB}{182,34,24}

\definecolor{lhc}{rgb}{1,0,0}   

\definecolor{lhc_1}{rgb}{1,0,0}
\definecolor{lhc_2}{rgb}{0,0,0}

\definecolor{midgreen}{HTML}{69c5a3}
\definecolor{midblue}{HTML}{6ca6cd}
\definecolor{darkgreen}{HTML}{77A351}
\definecolor{darkblue}{HTML}{74B8EF}
\definecolor{deepblue}{rgb}{0, 0, 0.55}
\definecolor{lightblue}{rgb}{0.68, 0.85, 0.9}

\definecolor{mygreen}{RGB}{93,173,85}

\definecolor{task1}{RGB}{225, 243, 240}
\definecolor{task2}{RGB}{230, 242, 255}

\usepackage{hyperref}
\hypersetup{colorlinks=true,citecolor=citecolor1,linkcolor=BrickRed,urlcolor=Thistle}

\definecolor{mygray}{gray}{.88}
\definecolor{mygrayd}{gray}{.94}
\definecolor{darkgreen}{HTML}{77A351}
\definecolor{darkblue}{HTML}{74B8EF}
\usepackage{tikz}

\def\eg{\emph{e.g.}}

\usepackage{mathrsfs}
\usepackage{multirow}
\usepackage{amsmath}
\usepackage{pifont}
\begin{document}

\title{VMAD: Visual-enhanced Multimodal Large Language Model for Zero-Shot Anomaly Detection}

\author{Huilin Deng, Hongchen Luo, Wei Zhai, Yang Cao, \IEEEmembership{Member, IEEE}, Yu Kang, \IEEEmembership{Senior Member, IEEE}
\thanks{Huilin Deng and Wei Zhai are with the School of Information Science and Technology, at the University of Science and Technology of China, Hefei, Anhui, China. (email: huilin\_deng@mail.ustc.edu.cn, wzhai056@ustc.edu.cn)}
\thanks{Hongchen Luo is at the Northeastern University, Shenyang, Liaoning, China. (email: luohongchen@ise.neu.edu.cn).}
\thanks{Yang Cao and Yu Kang are with the School of Information Science and Technology, at the University of Science and Technology of China, Anhui, China and also with the Institute of Artificial Intelligence, Hefei Comprehensive National Science Center. (email: \{forrest, kangduyu\}@ustc.edu.cn).}
\thanks{This work is supported by the National Natural Science Foundation of China (62033012 and 62306295).}
\thanks{Corresponding authors: Hongchen Luo, Yang Cao}}



\maketitle

\begin{abstract}

Zero-shot anomaly detection (ZSAD) recognizes and localizes anomalies in previously unseen objects by establishing feature mapping between textual prompts and inspection images, demonstrating excellent research value in flexible industrial manufacturing. However, existing ZSAD methods are limited by closed-world settings, struggling to unseen defects with predefined prompts. Recently, adapting Multimodal Large Language Models (MLLMs) for Industrial Anomaly Detection (IAD) presents a viable solution. Unlike fixed-prompt methods, MLLMs exhibit a generative paradigm with open-ended text interpretation, enabling more adaptive anomaly analysis. However, this adaption faces inherent challenges as anomalies often manifest in fine-grained regions and exhibit minimal visual discrepancies from normal samples. To address these challenges, we propose a novel framework VMAD (\textbf{V}isual-enhanced \textbf{M}LLM Anomaly Detection) that enhances MLLM with visual-based IAD knowledge and fine-grained perception, simultaneously providing precise detection and comprehensive analysis of anomalies. Specifically, we design a Defect-Sensitive Structure Learning scheme that transfers patch-similarities cues from visual branch to our MLLM for improved anomaly discrimination. Besides, we introduce a novel visual projector, Locality-enhanced Token Compression, which mines multi-level features in local contexts to enhance fine-grained detection. Furthermore, we introduce the Real Industrial Anomaly Detection (RIAD), a comprehensive IAD dataset with detailed anomaly descriptions and analyses, offering a valuable resource for MLLM-based IAD development. Extensive experiments on zero-shot benchmarks, including MVTec-AD, Visa, WFDD, and RIAD datasets, demonstrate our superior performance over state-of-the-art methods. The code and dataset will be available soon.

\end{abstract}

\begin{IEEEkeywords}
Anomaly Detection, Zero-Shot Learning, Multimodal Large Language Models.
\end{IEEEkeywords}

\section{Introduction} 

\begin{figure}[htpb] 
	\centering
	\includegraphics[width=0.96\linewidth]{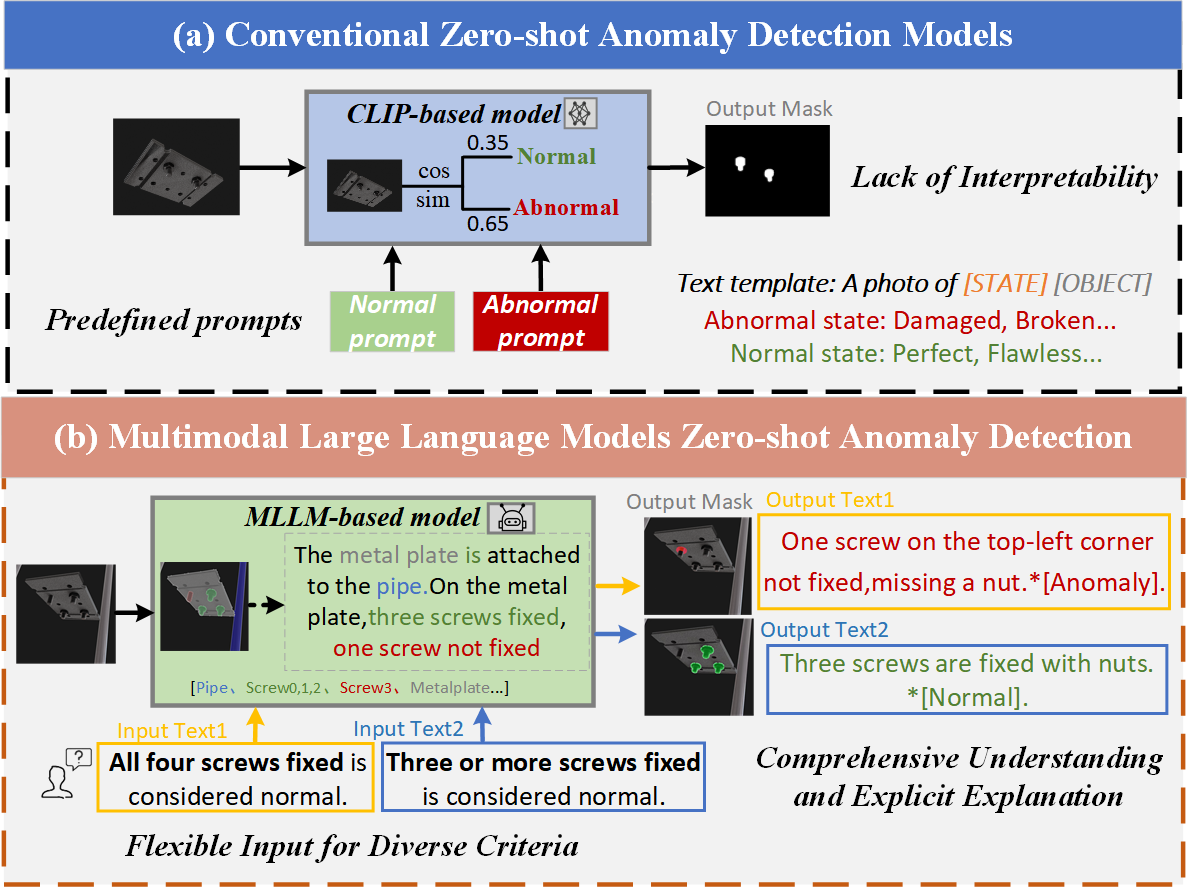}
	\caption{\textbf{Comparison between previous ZSAD methods and MLLMs-based ZSAD methods.} (a) Previous ZSAD methods use fixed templates and generic descriptions, confined to closed-world anomaly detection. (b) MLLMs-based methods leverage open-ended text interpretation and generation for IAD, providing additional comprehensive analysis and adapting flexibly to diverse criteria across multiple scenarios.}
 \label{Fig1}
\end{figure}

\IEEEPARstart{I}{ndustrial} Anomaly Detection (IAD) aims to classify and localize defects in industrial manufacturing.  By identifying abnormal patterns in industrial processes, IAD techniques enable timely intervention and optimization, thereby enhancing overall productivity. Collecting anomaly data is challenging due to their rarity and unpredictability \cite{yu2024tf}. Therefore, conventional IAD works \cite{liznerski2020explainable,li2021cutpaste,patchcore} mainly explore unsupervised techniques, measuring the deviation of test sample features from the learned normal distribution. However, these methods necessitate abundant training samples. Consequently, they exhibit limited generalization to novel classes and fail to adapt to dynamic production environments.

Recently, Zero-Shot Anomaly Detection (ZSAD) offers flexible inspection by using text prompts for anomaly measurement, enabling detection on unseen objects. Mainstream Contrast-based methods, based on pre-trained CLIP \cite{radford2021CLIP}, \textit{compares image features to textual descriptors representing `normal' and `abnormal' respectively}, as illustrated in Fig. \ref{Fig1} (a). WinCLIP \cite{jeong2023winclip} initially adapts CLIP using manually crafted prompts. AnomalyCLIP \cite{zhou2023anomalyclip} further substitutes manual templates with object-agnostic text vectors for generic representation. ClipSAM \cite{li2024clipsam} leverages SAM to CLIP's anomaly localization. Despite the promise, these methods still perform ZSAD in a \textbf{\textit{closed-world setting, executing binary classification in constrained semantic space with predefined prompts, thus struggling with unseen defects.}} Moreover, generic descriptors (\textbf{e.g.}, `damaged' in Fig.\ref{Fig1} (a)) are insufficient to capture diverse anomalies in industrial manufacturing.

Most recently, Multimodal Large Language Models (MLLMs) have emerged as a promising solution to the closed-world limitations. Unlike Contrast-based methods constrained in discriminative paradigm with predefined prompts, MLLMs exhibit a generative paradigm with open-ended text interpretation, enabling more adaptive anomaly analysis. Built upon large language models, MLLMs have demonstrated remarkable image-text understanding with more flexible input-output formats \cite{zhu2023minigpt,zhang2023llava}. This flexibility enables MLLMs to process arbitrary textual prompts with visual input, facilitating dynamic anomaly analysis based on various criteria and scenarios (as shown in Fig. \ref{Fig1} (b), the same sample yields different inspection results based on varying criteria). Therefore, adapting MLLMs for IAD tasks represents a promising yet underexplored field. Nevertheless, this adaptation faces inherent challenges. \textbf{\textit{Anomalies are visually confusing, characterized by minimal discriminative feature variances between normal and anomalous samples, mostly manifesting in object-localized regions.}} While MLLMs' general visual interpretation, they often struggle with abnormality discriminability and fine-grained perception of defects.

This weak discriminability often stems from minor manufacturing flaws or imperfections that blur the boundary between defective areas and surrounding normal regions. Intriguingly, our investigation uncovers a promising approach to address this issue. \textit{\textbf{Despite subtle global visual variances, normal and anomalous samples exhibit pronounced disparities in patch-similarity distributions. Specifically, normal patches frequently exhibit visual parallels across multiple unmarked samples, whereas anomalous areas rarely find such correspondences.}} Leveraging this patch-similarity disparity as valuable IAD-specific knowledge, we propose a \textbf{D}efect-\textbf{S}ensitive \textbf{S}tructure \textbf{L}earning (\textbf{DSSL}) scheme to amplify the abnormality distinction in LLM representation. Specifically, in the visual branch, DSSL computes local-global Visual Similarity between patch embeddings and global normal features. In LLM space, the similarity is calculated between visual-patch tokens and normal semantic tokens, which combines both linguistic and visual cues\cite{zhou2022MetaNet}. Finally, DSSL aligns these similarity distributions using a contrastive loss \cite{xuan2024decoupled}, effectively transferring patch-similarity cues to LLM space.

\begin{figure}[t] 
	\centering
	\includegraphics[width=0.9\linewidth]{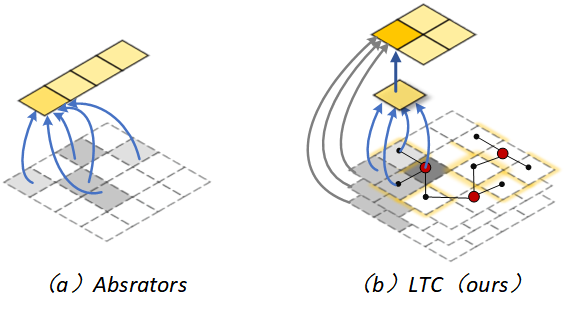}
 \vspace{-5pt}
	\caption{\textbf{Various visual projectors.} Abstractors compress limited information, while LTC mines multi-level local cues.
	}
	\label{projectors}
\end{figure}

The second challenge relates to fine-grained semantics of anomalies. We observe that visual projectors, responsible for transforming visual signals into LLM-compatible visual tokens, directly impact MLLM's perception of subtle anomalies. Traditional MLLMs' visual projectors use abstractors for efficiency \cite{meng2024deepstack,chen2024far,li2024tokenpacker}, but they compromise visual feature integrity by summarizing information from limited areas (Fig. \ref{projectors} (a)). To address this, we propose a \textbf{L}ocality-enhanced \textbf{T}oken \textbf{C}ompression (\textbf{LTC}) method which preserves rich semantics while reducing tokens. As outlined in Fig. \ref{projectors} (b), LTC mines multi-level features in local contexts by employing coarse-to-fine injection with multi-level features' integration. 

\begin{figure*}[t] 
	\centering
	\includegraphics[width=0.97\linewidth]{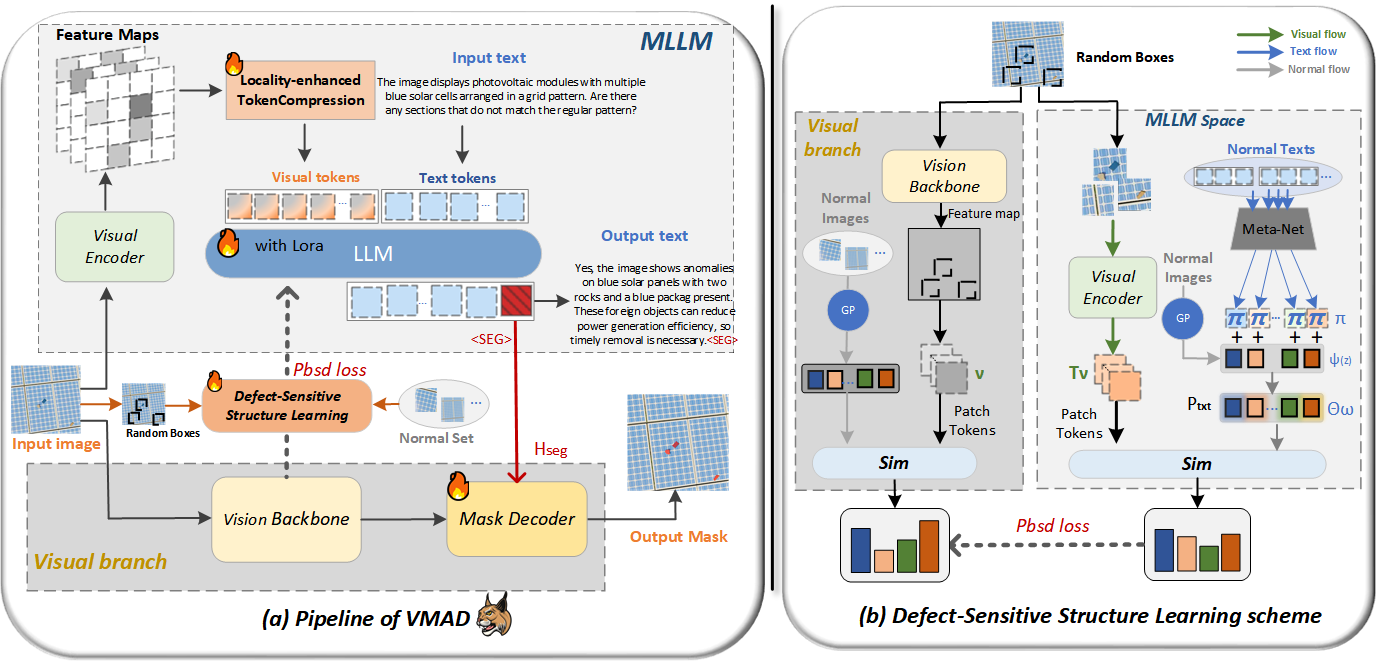}
	\caption{\textbf{(Left) Overview of VMAD.} VMAD incorporates a visual branch for anomaly localization (Sec. \ref{overview}), with Locality-enhanced Token Compression serving as a visual projector (Sec. \ref{LTC}). \textbf{(Right) Defect-Sensitive Structure Learning.} It aligns visual and text-visual patch similarity distributions using PBSD loss, enhancing MLLM's sensitivity to anomalous structures (Sec. \ref{PBSD}). \textbf{GP:} Global Pooling, \textbf{Sim:} Similarity computation.}
	\label{pipeline}
\end{figure*}

To this end, We propose VMAD (\textbf{V}isual-enhanced \textbf{M}LLM \textbf{A}nomaly \textbf{D}etection), a novel framework offering simultaneous anomaly localization and explainable text, with interactive engagement for follow-up inquiries. As illustrated in Fig. \ref{pipeline} (a), VMAD comprises an MLLM and visual branch. The MLLM processes image-text inputs, generating `[seg]' tokens for mask decoder's prompt segmentation. To address anomaly detection challenges, we introduce two innovative modules: 1) \textbf{DSSL}: Integrates visual-similarity cues across MLLM and visual branch for enhanced anomaly discrimination. 2) \textbf{LTC}: Processes visual embeddings within MLLM, producing high-quality tokens for fine-grained defect perception.

Moreover, we present \textbf{R}eal \textbf{I}ndustrial \textbf{A}nomaly \textbf{D}etection (\textbf{RIAD}), an extensive IAD dataset, encompassing 28,040 images across 24 object categories and 15 defect types. As shown in Fig. \ref{dataset}, RIAD features paired image-text data with anomaly masks, providing detailed anomaly descriptions, impact analyses, and recommendations, offering MLLMs crucial IAD-specific knowledge. Extensive experiments on zero-shot benchmarks, including MVTec-AD, VisA, WFDD, and our RIAD, demonstrate VMAD's superior performance over state-of-the-art methods. Furthermore, VMAD exhibits exceptional capability in providing accurate assessments and elaborative insights for industrial defects.

Our contributions are summarized as follows: \textbf{1)} We propose VMAD, a novel framework for IAD that simultaneously localizes anomalies and generates explanatory text. Additionally, we design a cross-modal learning scheme, DSSL, integrating visual similarity cues as IAD-specialized knowledge into MLLMs. \textbf{2)} We introduce a novel visual projector, the LTC mechanism, which mines multi-level features in local contexts to enhance fine-grained defect detection. \textbf{3)} We collect a dataset named RIAD containing plenty of paired image-text data with anomaly masks, providing a comprehensive resource for MLLM-based IAD development.

\section{Related Work}
\subsection{Industrial Anomaly Detection}
Given the scarcity of anomalies, conventional Industrial Anomaly Detection (IAD) research mainly explores unsupervised and self-supervised techniques. These approaches fall into two main streams. Reconstruction-based methods (\cite{surfaceAD_survey,reconstruction,jiang2022masked}) use encoder-decoder architectures to minimize input-reconstruction discrepancies. The Embedding-based branch (\cite{patchcore,deng2024PLMNet,chen2024unified}) identifies anomalies through feature disparities. It encompasses several sub-categories: a) One-class techniques(\cite{liznerski2020explainable,li2021cutpaste}). b) Memory-augmented models (\cite{patchcore,Graphcore}). c) Knowledge distillation frameworks (\cite{hinton2015distilling,zhai2023exploring}). Conventional methods follow `one-class-one-model' paradigm and struggle with novel object classes. In contrast, our VMAD enables in-context learning for multiple novel object categories.
\subsection{Zero-shot Anomaly Detection} 

Zero-shot anomaly detection \cite{jeong2023winclip,deng2023PTMNet,li2024clipsam}, using only text prompts even for unseen classes, has gained interest due to the increasing demand for flexibility in industrial settings. WinCLIP \cite{jeong2023winclip} pioneers language-driven zero-shot IAD using CLIP \cite{radford2021CLIP}, while AnoVL \cite{deng2023anovl} incorporates test-time adaptation for improved localization. As for another line, SAA \cite{cao2023segment} uses text prompts on SAM \cite{Kirillov_2023_ICCV} for candidate masks and filters them with complex mechanisms. Clipsam \cite{li2024clipsam} combines CLIP and SAM for segmentation and refinement. Despite strong zero-shot ability, they lack explainable analysis and depend on additional predefined steps like threshold filtering or manual prompts. Our proposed VMAD offers flexible detection and explainable analysis with a novel MLLM-based framework, 
\subsection{Multimodal Large Language Models (MLLMs)}
Large Language Models (LLMs) have shown remarkable capabilities in language tasks \cite{yin2023survey}. Building on this, Multimodal Large Language Models (MLLMs) extend these capabilities to the visual domain. Early efforts like Flamingo \cite{alayrac2022flamingo} and BLIP-2 \cite{li2023blip} pioneered the use of Q-Former and Resampler to bridge vision and language. Subsequent models such as LLaVA \cite{liu2024visual} and MiniGPT4 \cite{zhu2023minigpt} enhanced instruction following through visual instruction tuning. Recent advancements, including Ferret \cite{you2024ferret}, LISA \cite{lai2024lisa}, GLaMM \cite{rasheed2024glamm}, and LLaVA-Ground \cite{zhang2023llava}, have pushed MLLMs towards fine-grained visual understanding and grounded conversation generation.

In Industrial Anomaly Detection, AnomalyGPT \cite{gu2024anomalygpt} and Myriad \cite{li2023myriad} pioneer MLLM applications. To enable LLM's visual comprehension, AnomalyGPT converts anomaly maps to learnable embeddings while Myriad encodes anomaly maps into LLM-compatible tokens via Q-Former \cite{li2023blip}. However, both of them rely on pre-trained visual models for anomaly localization, confining LLMs to text responses and thus restricting overall generalization. Our VMAD extends LLMs to both anomaly localization and analysis, introducing a novel patch-based scheme for cross-modal generalization.

Visual projectors are crucial for real-time IAD tasks, efficiently transforming visual features for LLM compatibility. Existing methods like QPN \cite{yu2024QPN} and DeepStack \cite{meng2024deepstack} focus on query enhancement and token restructuring, while others use pixel shuffle \cite{chen2024far} or nearby concatenation \cite{dong2024internlm} to reduce visual tokens. However, these approaches compromise visual feature integrity in pursuit of efficiency \cite{li2024tokenpacker}. Our LTC mechanism incorporates multi-level visual cues through coarse-to-fine scheme, balancing efficiency with structural integrity.

\section{Method}
This section outlines the problem setup (Sec. \ref{Problem Setup}) and provides an overview of VMAD with its loss function (Sec. \ref{overview}). It then introduces the Defect-Sensitive Structure Learning scheme for MLLM and visual branch (Sec. \ref{PBSD}). Finally, we describe the Locality-enhanced Token Compression (LTC) mechanism as MLLM's visual projector (Sec. \ref{LTC}).

\subsection{Problem Setup} \label{Problem Setup}
We extend traditional mask-only zero-shot anomaly detection by incorporating text responses. This task aims to detect anomalies in novel object categories without support images while providing textual explanations. Training ($D_{train}$) and testing ($D_{test}$) sets contain disjoint object classes. Query images and text instructions are fed into the network in pairs. The text instruction $T_q$ includes \underline{\textit{a description of normal scenes}} and \underline{\textit{questions about anomalies}}. \textbf{\eg,\textit{`The gray metal plates in the picture should be secured with four pairs of screws, which must not be loose or missing. Is there any anomaly in this picture?'}} Given multi-modal inputs $(I_{img}, T_q)$ from $D_{train}$, the model learns to associate semantic information to provide text responses and segmentation masks. During testing, the model is evaluated without further optimization.

\subsection{Overall Architecture}\label{overview}
Our network consists of a standard MLLM and a visual branch. The Defect-Sensitive Structure Learning (DSSL) scheme spans both branches, using visual patch similarity to enhance MLLM's anomaly discrimination. The Locality-enhanced Token Compression (LTC) mechanism serves as the MLLM's visual projector, generating compact, detail-rich visual tokens. Subsequent sections elaborate on each module.

\begin{figure*}[htpb] 
	\centering
	\includegraphics[width=0.95\linewidth]{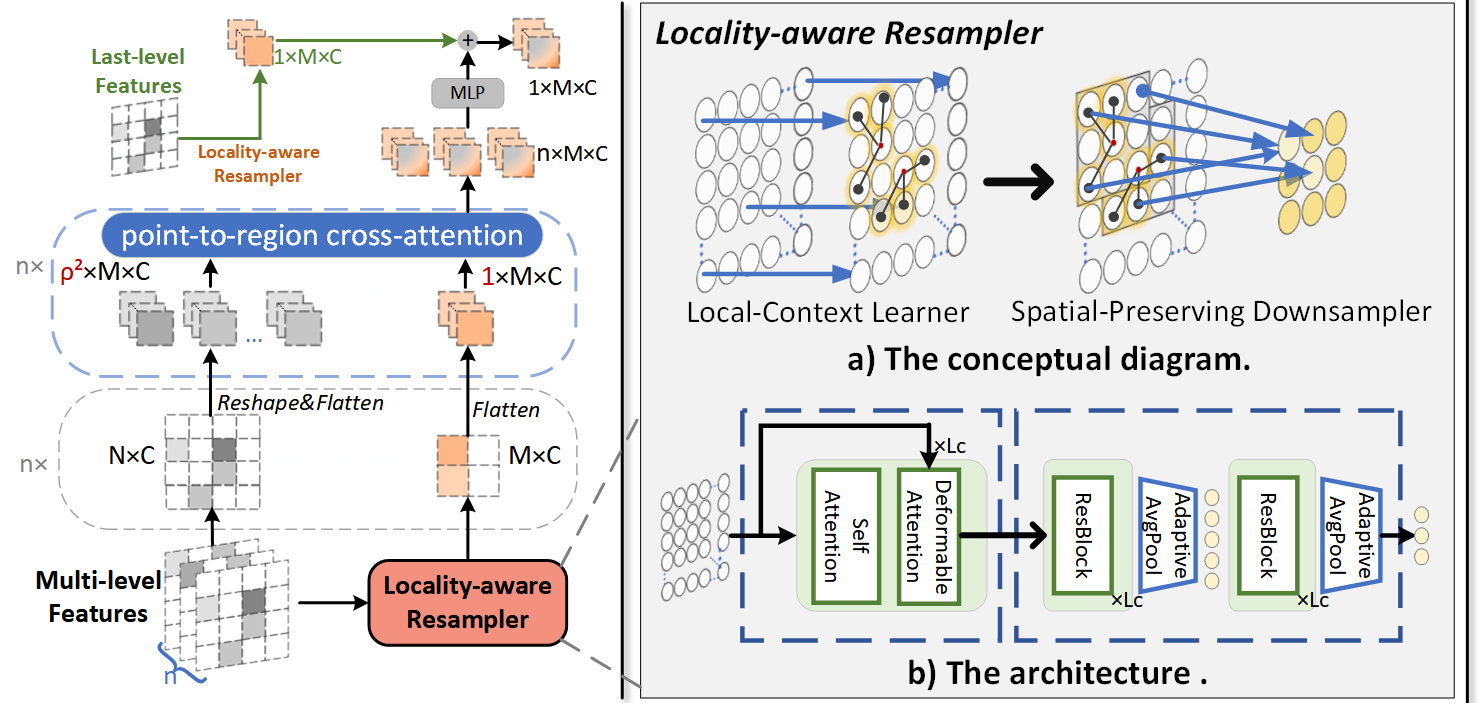}
	\caption{\textbf{Overview of LTC mechanism.} It incorporates multi-level visual cues through a coarse-to-fine scheme, providing comprehensive image information to the LLM.}  
	\label{Tockencompressor}
\end{figure*}

\subsubsection{MLLM Framework}
The proposed MLLM framework comprises three pivotal components: \textbf{(1) Visual Encoder $F_{I}$:}  the input image $I_{\text{img}} \in \mathbb{R}^{H \times W \times 3}$ is transformed into a set of visual embeddings $I_v \in \mathbb{R}^{N \times C}$ through the widely-utilized CLIP-ViT-L/14 \cite{radford2021learning} vision encoder. \textbf{(2) Visual Projector $\Psi_{I \rightarrow T}$:} it is responsible for projecting visual embeddings $I_v$ into visual token $T_v$ in the textual embedding space $T$, ensuring they have the appropriate dimension for subsequent language model. The visual projector takes the $N$ visual embeddings $I_v$ and converts them to $M$ visual tokens $T_v$, where $M < N$. \textbf{(3) LLM denoted as $\Theta(T_v, T_t)$:} it processes the visual token  $T_v$  and the textual token $T_t$, producing the coherent textual response auto-regressively. Formulated as:
\begin{equation}
\small
I_v=F_{I}(I_{\text{img}}),   
\end{equation}
\begin{equation}
\small
T_v=\Psi_{I \rightarrow T}(I_v),    
\end{equation}
\begin{equation}
\small
\hat{y}_{txt}=\Theta(T_v, T_t).   
\end{equation}

The embedding-as-mask scheme \cite{lai2024lisa} is leveraged to equip our MLLM with segmentation ability for anomaliy localization. In particular, we augment the LLM's vocabulary with a specialized $ < $seg$ > $ token,  which serves as the bridge between the MLLM and downstream mask decoder. Moreover, the text output $y_{txt}$ is ensured to end with the $ < $seg$ > $ token.

\subsubsection{Visual Branch}
The $ < $seg$ > $ token's last-layer embedding $\overline{H}_{seg}$ in the LLM is extracted and passed through an MLP projection layer $\Gamma(\cdot)$ to obtain $H_{seg}$. $H_{seg}$ is then used to inform the mask decoder about the semantic knowledge necessary for anomaly discrimination. $H_{seg}$ and $f$ are fed into the mask decoder to obtain the final segmentation mask, where the image $I_{\text{img}}$ are processed by the visual backbone (\eg, SAM \cite{Kirillov_2023_ICCV}) to extract visual feature maps $f$. The process can be expressed as:
\begin{equation}
\small
H_{seg}=\Gamma(\overline{H}_{seg}), f=V_{backbone}(I_{\text{img}}), 
\end{equation}
\begin{equation}
\small
\overline{M}=M_{decoder}(H_{seg},f).  
\end{equation}

\subsubsection{Training Objectives}
The model is optimized end-to-end through the text generation loss $\mathcal{L}_{txt}$, segmentation mask loss $\mathcal{L}_{seg}$ and the specially designed defect-structure loss $\mathcal{L}_{struc}$. The overall objective $\mathcal{L}$ is the weighted sum of these losses, formally defined as:
\begin{equation}
\small
\mathcal{L} = \lambda_{\text{txt}} \mathcal{L}_{\text{txt}} + \lambda_{\text{seg}} \mathcal{L}_{\text{seg}} + \lambda_{\text{pbsd}} \mathcal{L}_{\text{pbsd}}. 
\end{equation}

To encourage high-quality segmentation results, the segmentation loss $L_{seg}$ is computed as a combination of per-pixel binary cross-entropy (BCE) loss and DICE loss with corresponding loss weights $\lambda_{\text{bce}}$ and $\lambda_{\text{dice}}$. Concretely, given the predicted map $\hat{M}$ and query mask $M_q$, segmentation loss $\mathcal{L}_{seg}$ is formally defined as:
\begin{equation}
\small
\mathcal{L}_{\text{seg}} = \lambda_{\text{bce}} \textit{\textbf{BCE}}(\hat{M}, M_q) + \lambda_{\text{dice}} \textit{\textbf{DICE}}(\hat{M}, M_q).
\end{equation}

For text generation, $\mathcal{L}_{\text{txt}}$ is the auto-regressive cross-entropy loss between predicted text $\hat{y}_{txt}$ and the ground-truth text $y_{txt}$. The patch distribution similarity loss $\mathcal{L}_{pbsd}$ (Sec. \ref{PBSD} Eq.\ref{eq:pbsd}) aims to enhance anomalous structure discriminability by transferring patch-level similarity knowledge from visual domain to multimodal space, enabling consistent normal-abnormal patch distinction across modalities.

\subsection{Defect-Sensitive Structure Learning}\label{PBSD}
To enhance anomalous structure discriminability in LLM space, we introduce Defect-Sensitive Structure Learning (DSSL) scheme based on \textit{\underline{`Patch-Similarity'}}. DSSL calculates and aligns two similarity distributions: Visual Patch Similarity (between local patches and global features) and Text-Visual Patch Similarity (between visual patches and semantic tokens in LLM space). This alignment, achieved through the Patch-Based Similarity Distribution (PBSD) loss \cite{yu2023incremental}, transfers patch-level similarity knowledge from visual to LLM space, ensuring consistent normal-abnormal separation across modalities. For implementation, we randomly generate a set of patch boxes ${B_i[j]}_{j=1}^{N_b}$ from the input image $I{img}$, where $N_b$ is the number of patch boxes per image.

\subsubsection{Visual Patch Similarity Distribution} 
To obtain embedding features of patch boxes ${v[j]}_{j=1}^{N_b}$, we apply ROI pooling using the patch box coordinates, and then project the pooled features through a normalization head:
\begin{equation}
\small
f = V_{backbone}(I_{img}), v^j = g_{\alpha} (\text{ROI}(f; B[j])),
\end{equation}
where $g_{\alpha}$ represents projection head. Meanwhile, global normalized feature $z$ is obtained by applying global pooling to the feature map $f$, then projecting it through $g_{\beta}$: 
\begin{equation}
\small
z = g_{\beta}(GAP(h)).
\label{Eq:global_norm}
\end{equation}
Based on these global normalized features, we first establish a memory queue that contains all normal and abnormal samples from the same class as $I_{img}$:
\begin{equation}
\small
\text{M}_{\text{img}} = \{z_m: y_m=y^+ \text{ or } y_m=y^-\},
\end{equation}
where $y^+$ and $y^-$ denote the normal and abnormal class labels, respectively. We then define a positive set $\text{P}$ as a subset of $\text{M}$, \textit{containing only the global features of normal images from the same class as $I_{img}$.} The $\text{P}$ is formulated as:
\begin{equation}
\small
\text{P}_{\text{img}}=\{z_t: y_t=y^+\}.
\end{equation}

Finally, the similarity relationship between local patch and normal global representation is calculated as conditional probability, using a pre-defined temperature parameter $\tau$:
\begin{equation}
\small
p(\boldsymbol{z_t} | v^j) = \frac{\exp(\boldsymbol{z_t} \cdot \boldsymbol{v^j} / \tau)}{\sum_{\boldsymbol{z_m} \in \text{M}_{\text{img}}} \exp(\boldsymbol{z_m} \cdot v^j / \tau)}
\label{sim_vis}
\end{equation}
Eq.\ref{sim_vis} encodes similarity between local patch ($v^j$) and global representations ($z_t$). High similarity indicates normal patterns, while low similarity suggests anomalies. This local-global distinction is then used to guide normal-abnormal differentiation in LLM space, enhancing LLM's anomaly discrimination.

\subsubsection{Text-Visual Patch Similarity Distribution}\label{t-v_branch} It calculates the similarity between visual-patch tokens and semantic tokens of normal samples in LLM space. Notably, semantic tokens combine both linguistic and visual cues for a comprehensive representation of normality. To obtain patch tokens $\{T_v^j\}_{j=1}^{N_b}$ in LLM space, we extract image patches $\{I_{box}^j\}_{j=1}^{N_b}$ from the original image and project them into the LLM space:
\begin{equation}
\small
I_{box}^j=\text{ROI}(I_{img}; B[j]), T_v^j=\boldsymbol{\Psi_{I \rightarrow T}}(F_{I}(I_{box}^j)). 
\end{equation}
To obtain multimodal semantic tokens for image $x$, we generate text-based tokens $\pi$ by passing normal textual descriptions through Meta-Net \cite{zhou2022MetaNet}. We then enrich these linguistic representations with structural details by projecting normalized features $z$ of normal images into LLM space. Finally, the text-based tokens $\pi$ and projected global visual tokens are combined through a simple additive operation. The multimodal semantic tokens are calculated as:
\begin{equation}
\small
\boldsymbol{\theta_{\omega}(x)}= \boldsymbol{\Psi_{I \rightarrow T}}(z)+\pi,\quad \pi=\boldsymbol{Meta}(t_x),
\label{semantic_tokens}
\end{equation}
where $\boldsymbol{\theta_{\omega}(x)}$ represents multimodal semantic tokens for the image $x$. The global normalized feature $z$ is obtained consistent with our visual branch (Eq.\ref{Eq:global_norm}), while the feature is extracted through MLLM's visual encoder. $\boldsymbol{\Psi_{I \rightarrow T}}$ represents the projection of global normalized feature from visual to LLM space. In this work, the Meta-Net is built with a two-layer bottleneck (Linear-ReLU-Linear). Based on Eq. \ref{semantic_tokens}, we then define the positive set $\text{P}_{\text{txt}}$ and memory queue $\text{M}_{\text{txt}}$ as follows:
\begin{equation}
\small
\text{P}_{\text{txt}}=\{ \boldsymbol{\theta_{\omega}}(x_t): y_t=y^+ \},
\end{equation}
\begin{equation}
\small
\text{M}_{\text{txt}} = \{\boldsymbol{\theta_{\omega}}(x_m): y_m=y^+ \text{ or } y_m=y^-\}.
\end{equation}

Similar to Eq. \ref{sim_vis}, the similarity distribution between patch tokens $T_v^j$ and normal semantic tokens $\boldsymbol{\theta_{\omega}(z)}$ is computed as:
\begin{equation}
\small
p(\boldsymbol{\theta_{\omega}}^t | T_v^j) = \frac{\exp(\boldsymbol{\theta_{\omega}}^t \cdot \boldsymbol{T_v}^j / \tau)}{\sum_{\boldsymbol{\theta_{\omega}^m} \in \text{M}_{\text{txt}}} \exp(\boldsymbol{\theta_{\omega}}^m  \cdot v^j / \tau)}.
\label{sim_LLM}
\end{equation}

To align local-global visual distributions and multimodal semantic distributions obtained by Eq. \ref{sim_vis} and Eq. \ref{sim_LLM}, respectively, we introduce the Patch-Based Similarity Distribution (PBSD) loss. Mathematically, PBSD loss can be expressed as:
\begin{equation}
\small
\mathcal{L}_{pbsd} = \frac{1}{N_b} \sum_{j=1}^{N_b} \sum_{\boldsymbol{x_t} \in \text{P}} -p(\boldsymbol{z_t} | v^j) \log p(\boldsymbol{\theta_{\omega}}^t | T_v^j),
\label{eq:pbsd}
\end{equation}
where $p(\boldsymbol{z_t} | v^j) $ is detached from the computation graph to prevent gradient flow. The PBSD loss effectively transfers the knowledge of patch-level similarities from visual domain to multimodal space, ensuring the model learns to distinguish normal and abnormal patches consistently across modalities.

\begin{figure*}[t] 
	\centering
	\includegraphics[width=0.99\linewidth]{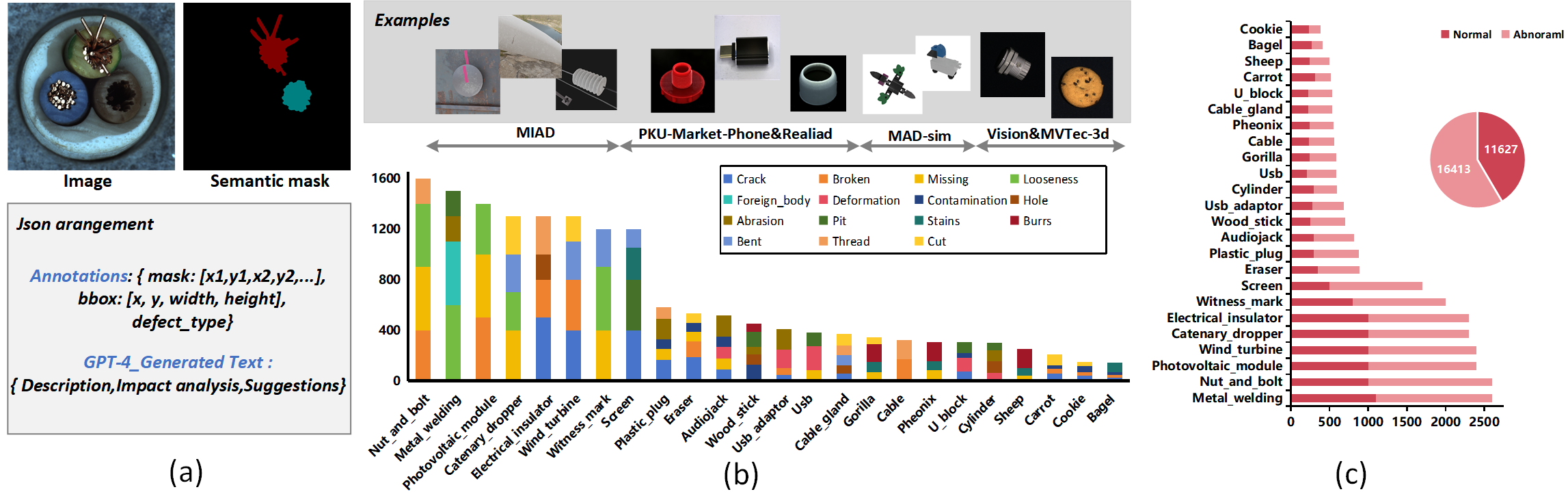}
	\caption{\textbf{Properties of the RIAD dataset.} (a) RIAD data pairs: images, semantic segmentation masks encoding defect types, and GPT-generated text. Masks and text are stored in JSON format. (b) The horizontal axis represents the categories of objects, the vertical axis represents quantity, and different colors represent different defect types. The top part shows sample RIAD images with their source datasets indicated. (c) The ratio of normal images and abnormal images in each object class.}
 \label{dataset}
\end{figure*}

\subsection{Locality-enhanced Token Compression}\label{LTC} 
In real-time industrial inspection, abstractors are preferred for efficiency but struggle with spatial information (Fig. \ref{projectors} (a)). Our Locality-enhanced Compression balances spatial details and efficiency by mining multi-level features in local contexts (Fig. \ref{projectors} (b)). It performs Locality-enhanced downsampling with a dual-faceted refinement, comprising Coarse-to-fine Refinement and Multi-level Feature Integration (Fig. \ref{Tockencompressor}).

\noindent\textbf{\emph{Locality-enhanced downsampling.}} We downsample MLLM's last-layer visual features $I_v\in \mathbb{R}^{N \times C}$ to $I_v^{\downarrow} \in \mathbb{R}^{M \times C}$ with our locality-aware resampler, where $\rho$ is downsampling rate. It combines convolution and deformable attention to handle diverse anomaly shapes and sizes. As shown in Fig.\ref{Tockencompressor} (right) (a), it comprises a \noindent\underline{\textit{Local-Context Learner}} to enhance locality awareness while maintaining flexibility towards geometric variations, and a \noindent\underline{\textit{Spatial-Preserving Downsampler}} to retain local context while abstracting features. The Learner adapts to geometric variations with flexible sampling locations. Each learnable query gathers visual cues through a 2D coordinate-based sampling using reference points and sampling offsets:
\begin{equation} 
\small
p' = p + \Delta p,
\end{equation}
\begin{equation} 
\small
O(i, j) = \sum_{k=0}^{K-1} W_k * X(i + s*(p' + \Delta p), j),
\end{equation}
where $s$ denotes stride and the learned offset $\Delta p$ is added to sampling locations. Fig. \ref{Tockencompressor} (right) (b) illustrates the structure of locality-aware resampler. Local-Context Learner stacks $\mathcal{L}_c$ blocks, each containing self-attention and deformable attention layers. Spatial-preserving downsampler comprises $\mathcal{L}_c$ ResNet\cite{resnet} blocks, followed by adaptive average pooling, and another $\mathcal{L}_c$ ResNet blocks. This design balances locality-awareness with efficient feature abstraction.

\noindent\textbf{\emph{Coarse-to-fine Refinement.}} The refinement process begins with coarse-to-fine mappings, integrating detailed cues from high-resolution features into the downsampled coarse representation. As the coarse representation $I_v^{\downarrow}$, \textbf{\textit{each pixel in $\boldsymbol{I_v^{\downarrow}} \in \boldsymbol{\mathbb{R}^{1 \times M \times C}}$ corresponds to $\boldsymbol{\rho \times \rho}$ sub-region in $\boldsymbol{I_v}\in \boldsymbol{\mathbb{R}^{ \rho^2 \times M \times C}}$}}. This mechanism enables low-resolution queries to assimilate fine-grained keys and values. Specifically, we consider the low-resolution$I_v^{\downarrow} \in \mathbb{R}^{1 \times M \times C}$ as point-based queries while $I_v$ as region-based keys and values. For level $l$, we compute the attention scores:        
\begin{equation}
\small
V_{att,l}=softmax(\frac{Q^{\downarrow}_lK^T_l}{\sqrt{d}})V_l,
\end{equation}
where $Q^{\downarrow}_l$ represents the queries derived from coarse features, $K_l, V_l$ are keys and values from the original ones.

\noindent\textbf{\emph{Multi-level Feature Integration.}} To achieve hierarchical semantic representations, multi-level visual features are integrated as enriched reference keys and values. This leverages inherent biases of different CLIP encoder layers towards distinct visual patterns: shallow layers capture low-level details, while deeper layers encode semantic understanding. Concretely, the aggregation values incorporate the past $n$ levels:
\begin{equation}
\small
V_{att,all}=Linear(cat([V_{att,L-n-1},...V_{att,L-2}])),
\label{multi_layer}
\end{equation}
where $L$ is the total levels in CLIP encoder. Finally, we enhance last-layer coarse representation with aggregated values:
\begin{equation}
\small
I^{*}_v= I_v^{\downarrow}+V_{att,all}.   
\end{equation}
LTC maintains rich semantic cues while reducing visual tokens, simultaneously performing coarse-to-fine injection and multi-level feature integration.

\section{Dataset}\label{novel_dataset}

\subsection{Collection Details.}
We collect the Real Industrial Anomaly Detection Dataset (RIAD), extracted from various practical industrial datasets including MIAD \cite{bao2023miad}, Realiad \cite{wang2024realiad}, Vision \cite{bai2023vision}, MAD-sim \cite{zhou2024pad}, MVTec-3D \cite{bergmann2021mvtec3d}, and PKU-market-iphone \cite{zhang2022iphonescreen}. RIAD pairs each image with an anomaly mask and text data (Fig. \ref{dataset} (a)). As illustrated in Fig. \ref{dataset} (b) (top), RIAD encompasses a rich variety of industrial scenarios, including outdoor environments such as wind power equipment, indoor industrial components, and food products. This diverse range provides a more comprehensive representation of real-world industrial defects. The RIAD contains 11,627 normal and 16,413 defect images across 24 object classes and 15 defect types, providing extensive resources for IAD model development.

\subsection{Annotation Details.}
We utilize Anylabeling to label defect types within the anomaly regions. These semantic masks are recorded in COCO format with defect-category IDs ranging from 0 to 14, encompassing 15 distinct anomaly categories. Notably, we've transformed original binary (normal/abnormal) masks into semantic masks that distinguish different defect types, while also refining low-quality defect masks for improved accuracy. Meanwhile, images are also annotated with bounding boxes of anomaly regions. To provide rich, contextual textual data for each anomaly instance, we feed the anomaly images and COCO-format masks to GPT-4\cite{chen2024far}, generating three types of question-answer pairs: anomaly descriptions, impact analyses, and suggestions. This comprehensive textual information distinguishes RIAD from existing datasets, offering a holistic resource for IAD models leveraging both visual and textual inputs. Paired examples are shown in Fig. \ref{dataset} (a).

\subsection{Statistic Analysis.}
Fig. \ref{dataset} (b) shows the count and distribution of anomalies in images The top of Fig. \ref{dataset} (b) displays image samples, showcasing diverse scenarios from outdoor industrial settings to indoor electronics, ceramics, building blocks, and food items. Fig. \ref{dataset} (c) illustrates the ratio of normal images and abnormal images in each object category. In the cross-category setting, we train on 16 classes (15,274 images) and test on 8 unseen classes (8,342 images) during training.

\begin{table*}[t]
	\centering
	\caption{\textbf{Zero-shot Anomaly Detection Performance on \colorbox{task1}{Cross-dataset Evaluation} setting on MVTec-AD, Visa, and WFDD datasets.} The best results are in \textbf{bold} while the second best are underlined. `Img' and `Pixel' represent the mean image-level and pixel-level AUROC.}
	\renewcommand{\arraystretch}{1.}
	\renewcommand{\tabcolsep}{8. pt}
	\begin{tabular}{l||cccc|cccc|cccc}
		\Xhline{2.\arrayrulewidth}
		\multirow{2}{*}{\textbf{Methods}} & \multicolumn{4}{c|}{\textbf{MVTec-AD}} & \multicolumn{4}{c|}{\textbf{ViSA}} & \multicolumn{4}{c}{\textbf{WFDD}}  \\
		\cline{2-13}
		& Img & Pixel & PRO & AP & Img & Pixel & PRO & AP & Img & Pixel & PRO & AP  \\
		\hline
		LISA \cite{lai2024lisa} & 89.1 & 91.6 & 82.2 & 81.2 & 81.3 & 85.2 & 77.8 & 79.3 & 92.1 & 93.5 & 86.0 & 87.2 \\
		BGAD \cite{BGAD} & 90.1 & 91.3 & 87.3 & 85.6& $\underline{87.4}$& 88.6& 82.0 & 80.6& 95.0 & 96.2 & 86.8 & 88.1  \\
		AnomalyClip \cite{zhou2023anomalyclip} & 91.5 & $\underline{94.7}$ & 86.2 & 83.4 & 82.3 & 87.1 & $\underline{84.2}$ & 80.3& 94.1 & 95.3 &84.9 & 83.8 \\
		VAND \cite{chen2023zero} & 91.8 & 92.5 & 88.2 & 86.8 & 83.5 & 90.7 & 81.9 & 82.8 & $\underline{94.6}$ & 96.1 & 84.3 & $\underline{88.5}$\\
		WinClip \cite{jeong2023winclip} & $\underline{92.5}$ & 93.4 & $\underline{89.5}$ & 87.5 & 84.8 & 91.3 & 80.4 & 82.4 & 93.2 & 96.2 & 86.1  & 83.5 \\
		AnomalyGPT \cite{gu2024anomalygpt} & 92.1 & 93.9 & 86.2 & 85.1 & 86.6 & $\underline{92.4}$ & 81.7 & 75.6 & 93.7 & \textbf{96.9}  & 85.7 & 84.2\\
		Myriad\cite{li2023myriad} & 93.2 & 92.3 & 87.9  & \textbf{88.2}  & 85.9  & 90.5 & 81.3 & 83.5  & 91.8 & 93.9 & $\underline{87.3}$ & 82.7 \\
		\hline
		\rowcolor{mygray}
		VMAD & \textbf{95.8} & \textbf{96.1} & \textbf{91.2} & $\underline{87.6}$ & \textbf{89.7} & \textbf{93.8} & \textbf{84.6} & \textbf{85.2} & \textbf{95.3} & $\underline{96.7}$  &\textbf{89.2} & \textbf{90.1} \\
		\bottomrule
	\end{tabular}
	
	\label{tab_crossdataset}
\end{table*}

\begin{figure*}[htpb] 
	\centering
	\adjustbox{max width=0.9\linewidth, max height=0.8\textheight}{%
		\includegraphics{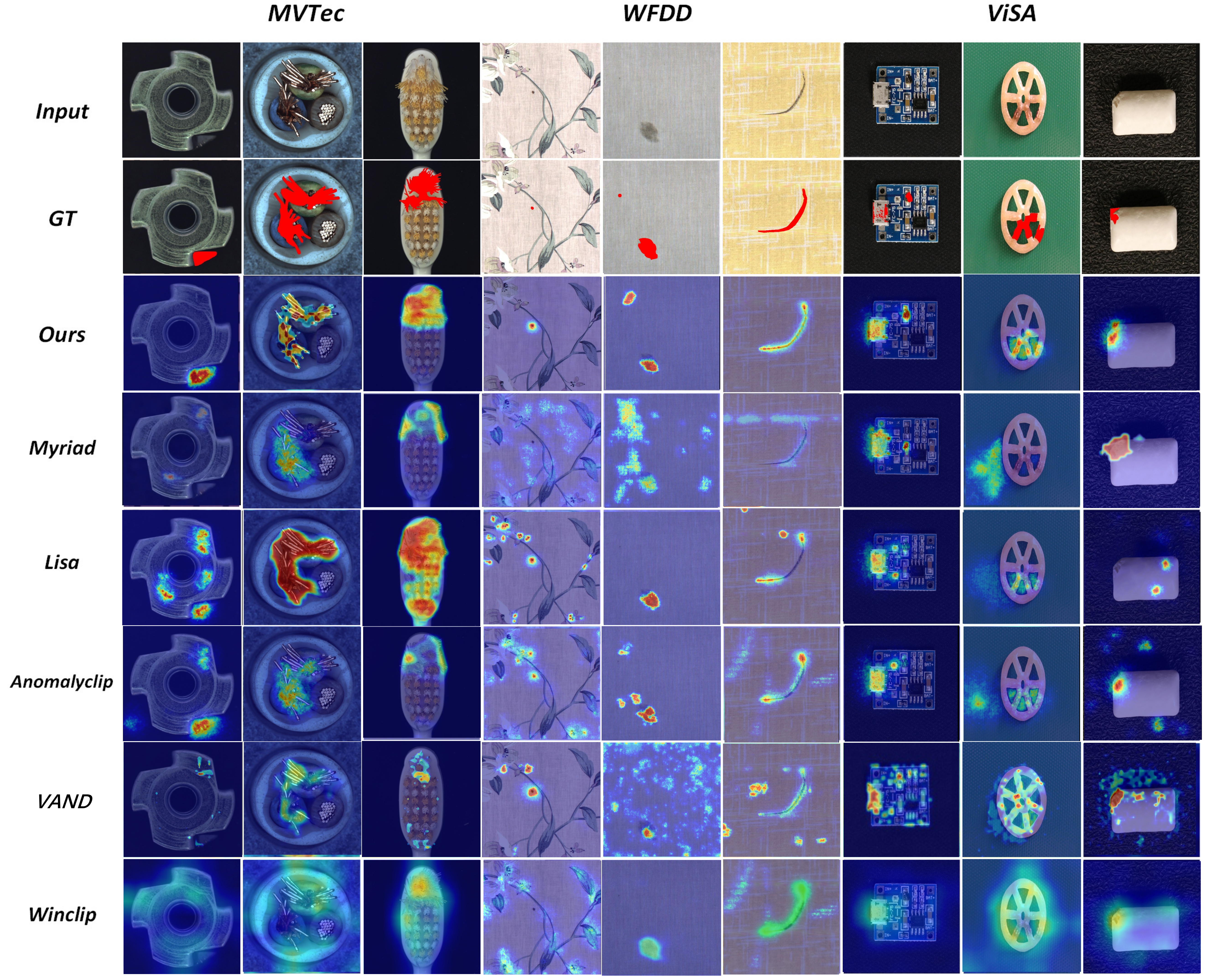}
	}
	\caption{\textbf{Zero-shot anomaly segmentation on MVTec-AD, WFDD, and ViSA datasets.}}
	\label{fig:vis_concat}
\end{figure*}

\section{Experiments}\label{exp} 
This section elaborates on the experiments' details. Section \ref{settings} presents the experimental protocol. Section \ref{results} analyzes the main results. Section \ref{ablation} demonstrates the ablation study.

\subsection{Experimental Protocol}
\subsubsection{Evaluation Protocols}\label{settings}
We evaluate models' adaptability to unseen scenarios using two complementary settings:
\begin{itemize}
	\item{\noindent\underline{\textit{Cross-dataset Evaluation:}}} Following \cite{gu2024anomalygpt,li2023myriad}, we adopt cross-dataset zero-shot anomaly detection, training on RIAD, and evaluating three widely-used public datasets without fine-tuning: \textbf{MVTec-AD} \cite{mvtec}: 5,354 high-resolution images across 15 classes.\textbf{Visa} \cite{zou2022spot}: The largest industrial anomaly detection dataset, with 10,821 images across 12 categories of colored industrial parts. \textbf{WFDD} \cite{chen2024unified}: 4,101 images of woven fabrics with different textures and patterns across 4 categories. 
	\item{\noindent\underline{\textit{Cross-category Evaluation:}} We randomly split RIAD into non-overlapping subsets: 16 classes ((15,274 images)) for training and 8 classes (8,342 images)  for testing.}
\end{itemize}

\begin{table*}[h]
	\centering
	\caption{\textbf{Zero-shot Anomaly Detection Performance on \colorbox{task2}{Cross-category Evaluation} across each class in RIAD.}}
	\renewcommand{\arraystretch}{1.}
	\renewcommand{\tabcolsep}{5.5 pt}
	\begin{tabular}{@{}l||ccccccccccccccccc@{}}
		\toprule
		\multirow{2}{*}{\textbf{Methods}} & \multicolumn{3}{c}{\textbf{metal\_welding}} & \multicolumn{3}{c}{\textbf{u\_block}} & \multicolumn{3}{c}{\textbf{toy\_brick}} &...& \multicolumn{3}{c}{\textbf{wind\_turbine}} & \multicolumn{3}{c}{\textbf{Average}} \\
		\cline{2-17}
		& Img & Pixel & PRO & Img & Pixel & PRO & Img & Pixel & PRO & & Img & Pixel & PRO & Img & Pixel & PRO \\
		\hline
		LISA \cite{lai2024lisa} & 84.1 & 90.5 & 81.0 & 89.1 & 91.2 & 86.6 & 80.4 & 82.3 & 76.1 & & 81.8 & 89.5 & 82.7 & 79.2 & 82.7 & 78.9 \\
		BGAD \cite{BGAD} & 91.6 & 94.7 & 87.2 & 79.1 & 82.4 & 77.2 & 86.2 & 87.3 & 85.2 & & 91.8 & 93.5 & 87.1 & 89.2 & 91.4 & 87.1 \\
		AnomalyClip \cite{zhou2023anomalyclip} & 90.2 & 92.2 & $\underline{91.1}$ & 87.4 & 90.7 & 89.2 & 89.1 & 93.7 &  89.2 &... & 88.9 & 90.6 & 89.1 & 92.7 & $\underline{96.6}$ & $\underline{90.5}$ \\
		VAND \cite{chen2023zero} &  $\underline{96.2}$ &  95.4 & 94.2 & 91.2 & \underline{96.3} & \underline{89.7} & $\underline{91.7}$ & 92.3 & 86.9 & & 78.8 & 82.6 & 88.1 & 92.1 & 94.7 & 89.4 \\
		WinClip \cite{jeong2023winclip} & 94.8 & $\underline{96.1}$ & 90.8 & $\underline{90.8}$ & 95.2 & 87.6 & 90.4 & 91.5 & 85.3 & & 89.2 & \underline{94.6} & \underline{87.9} & \underline{94.2} & 93.1 & 86.1 \\
		AnomalyGPT\cite{gu2024anomalygpt}& 90.1 & 86.2 & 84.6 & 83.7 & 87.1 & 88.2 & 89.5 & 92.4 &  80.1 & & 78.2 & 78.1 & 67.8 & 87.9 & 89.4 & 87.1\\
		Myriad \cite{li2023myriad} & 95.2 & 93.1  & 88.1 & 84.6 & 84.4 & 89.4 & 93.3 &  $\underline{95.8}$ & 90.8 & & 89.1 & 81.9 & 87.4 & 91.1 & 92.4 & 85.2 \\
		\hline
		\rowcolor{mygray}
		VMAD & \textbf{97.2} & \textbf{98.6} & \textbf{91.6} & \textbf{91.1} & \textbf{98.9} & \textbf{90.2} & \textbf{97.2} & \textbf{98.1} & \textbf{92.3} & &\textbf{93.7} & \textbf{98.1} & \textbf{89.5} & \textbf{94.9} & \textbf{98.9} & \textbf{92.3} \\
		\bottomrule
	\end{tabular}%
	\label{tab:RIAD}
\end{table*}

\subsubsection{Training Data Formulation} 
Our multi-modal training approach incorporates three distinct task types:
\begin{itemize}
	\item{\noindent\underline{\textit{Anomaly Segmentation Task.}}} The model outputs only segmentation mask. We employ a question-answer template like "USER: $ < $IMAGE$ > $ Are there any abnormalities present in the {OBJECT}? If so, please output the defect segmentation result. ASSISTANT: It is $ < $SEG$ > $."
	\item{\noindent\underline{\textit{Anomaly Segmentation and Answering Task.}}} The model produces both anomaly mask and corresponding textual answers. We employ a question-answer template: ``USER: $ < $IMAGE$ > $ Are there any abnormalities present in the {OBJECT}? If so, please output the defect segmentation result and provide QUERY TYPE. ASSISTANT: It is $ < $SEG$ > $. {ANSWERING}'' Queries fall into three types: description, impact analysis, and suggestion analysis, aiming to provide comprehensive insights for anomalies. The example responses are illustrated in Fig.\ref{chat}.
	\item{\noindent\underline{\textit{Visual Question Answering Task.}}} To preserve the original Visual Question-answering ability of LLM, we include the VQA task during training. We generate question-answer pairs based on defect descriptions using GPT-4. 
\end{itemize}

\subsubsection{Baselines}\label{Baselines}
For a comprehensive evaluation, we compare PLMNet against 7 typical baseline methods: two MLLM-based anomaly detection models (\textbf{Myriad} \cite{li2023myriad} and \textbf{AnomalyGPT} \cite{gu2024anomalygpt}), three zero-shot anomaly detection models (\textbf{WinClip} \cite{jeong2023winclip}, \textbf{VAND} \cite{chen2023zero} and \textbf{AnomalyClip} \cite{zhou2023anomalyclip}), and one supervised anomaly detection model (\textbf{BGAD} \cite{BGAD}) and one multi-modal pixel-grounding model (\textbf{LISA} \cite{lai2024lisa}). The anomaly detection performance is evaluated using the \textbf{A}rea \textbf{U}nder the \textbf{R}eceiver \textbf{O}perating \textbf{C}haracteristic Curve (\textbf{AUROC}). Additionally, \textbf{A}verage \textbf{P}recision (\textbf{AP}) for \textbf{Im}a\textbf{g}e-level detection (\textbf{Img}) and AUPRO for pixel-level segmentation are also included to provide more in-depth analysis.

\subsubsection{Implementations}\label{Implementation} We employ LLaVA-7B-v1-1 \cite{liu2024visual} as our multi-modal LLM and utilize the ViT-H SAM for visual backbone. We keep the vision encoders and visual backbone frozen. Concurrently, the LLM is fine-tuned using the LoRA \cite{hu2021lora} technique. The memory bank (Sec. \ref{PBSD}) contains $1/20$ of both normal and anomalous samples from each class. We adjust the down-sampling ratio $\rho=2$ in Sec. \ref{LTC}, and set $n=4$ in Eq. \ref{multi_layer}.  Models are trained for 10 epochs using an AdamW optimizer with cosine learning rate schedule on a 4×4090 GPU cluster, processing two input pairs per iteration.

\begin{figure*}[htpb] 
	\centering
	\includegraphics[width=0.93\linewidth]{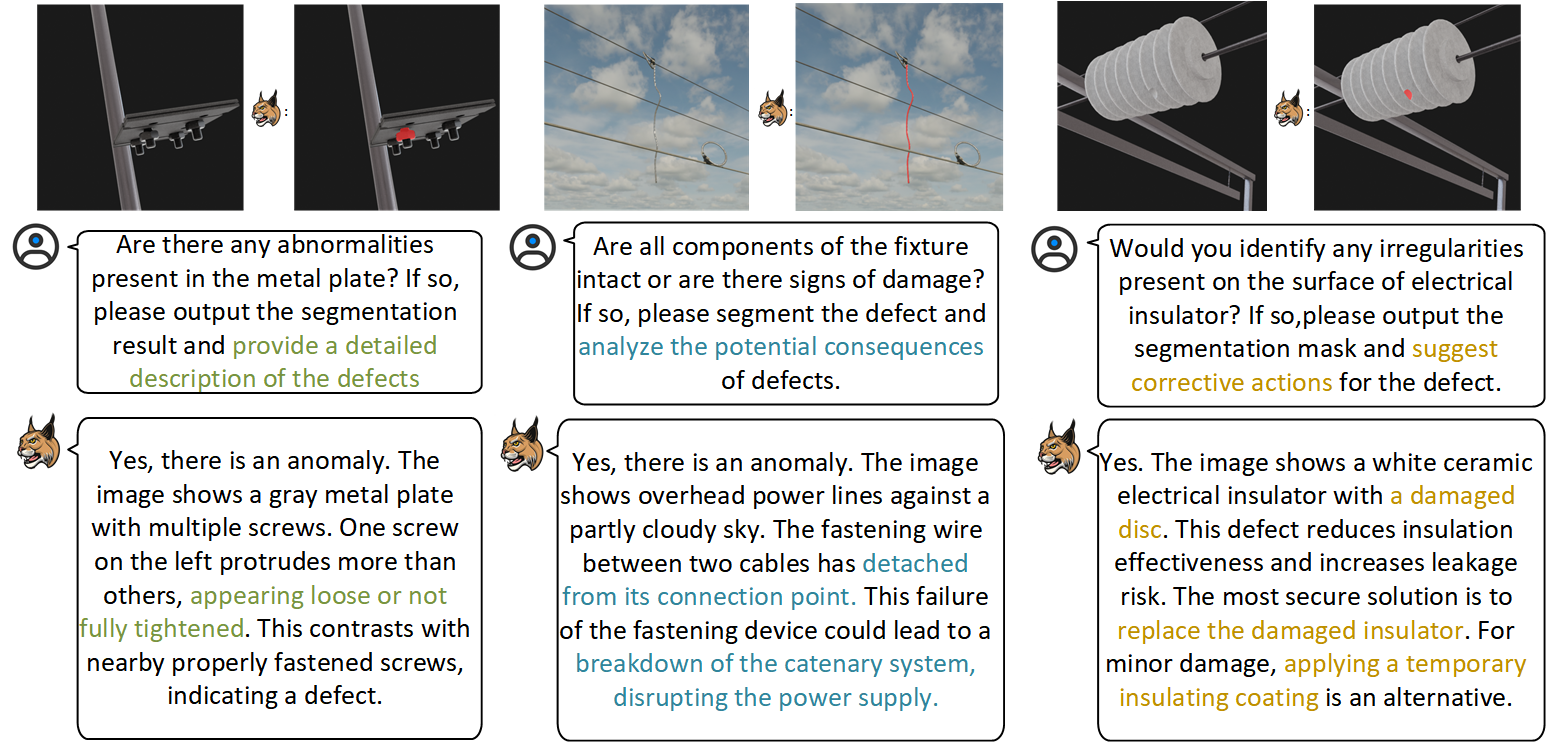}
	\caption{\textbf{Qualitative examples of VMAD in cross-category setting.} VMAD provides pixel-level anomaly localization and answers inspection-related questions, focusing on image description, anomaly impact, and suggestions.}
	\label{chat}
\end{figure*}

\subsection{Results and analysis}\label{results}
In this section, we compare VMAD with previous IAD works through numerical analysis and visual examples.

\noindent\textbf{\emph{Cross-dataset Evaluation:}}\label{results_MVTec} Table \ref{tab_crossdataset} summarizes the zero-shot results on MVTec-AD, Visa, and WFDD datasets. Tab. \ref{tab_crossdataset} reveals that VMAD outperforms competing methods in the presence of large domain gaps, particularly excels in pixel-level and region-level analyses, with margins of \textbf{3.3\%}p and \textbf{3.8\%}p compared to Myriad\cite{li2023myriad}. This might be attributed to keen visual feature-capturing and sensitivity to anomalous structures. Fig. \ref{fig:vis_concat} visually displays the anomaly localization results, highlighting VMAD's improved segmentation precision. 

\begin{figure}[!t] 
	\centering
	\includegraphics[width=0.89\linewidth]{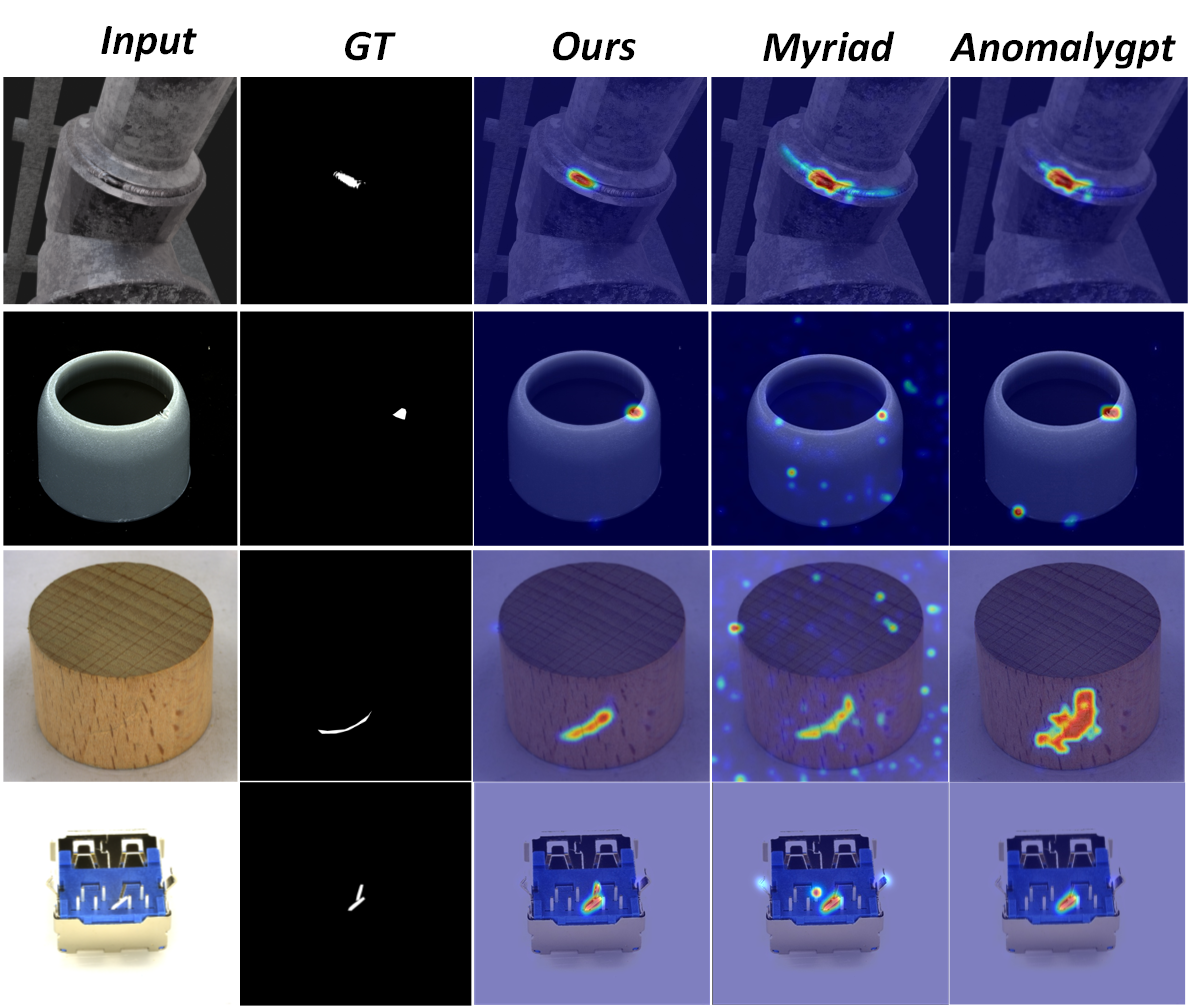}
	\caption{\textbf{Comparative visualization of zero-shot anomaly segmentation on RIAD dataset.}  
	}
	\label{vis_RIAD}
\end{figure}

\noindent\textbf{\emph{Cross-category Evaluation:}}\label{results_RIAD} Tab. \ref{tab:RIAD} details the performance, demonstrating state-of-the-art results of VMAD. Notably, our model particularly excels in pixel-level analyses, achieving a substantial \textbf{2.0\%}$\sim$\textbf{14.5\%} AUC-PR improvement over other multimodal IAD methods. This underscores VMAD's exceptional capability in anomaly localization. Fig. \ref{vis_RIAD} shows visualization results, while Fig. \ref{chat} displays VMAD's dialogue and segmentation on RIAD. Beyond precise localization, VMAD demonstrates remarkable zero-shot understanding and reasoning in anomaly detection, offering potential for production guidance through impact analysis and suggestions.

\section{Ablation Study}\label{ablation}
In this section, we conduct a comprehensive ablation study to validate the contributions of major modules. All experiments are performed under two settings: cross-dataset setting on MVTec-AD or cross-category setting on RIAD.

\noindent\textbf{\emph{Ablation Analysis of Main Components:}} We report the experimental results of selectively activating modules in Tab. \ref{tab_ablation}. It reveals that activating DSSL alone improves RIAD's AUROC by \textbf{3.1\%}p (image-level) and \textbf{3.0\%}p (pixel-level). LTC enhances MVTec's performance by \textbf{2.7\%}p (pixel-level AUROC) and \textbf{2.0\%}p (region-level AURPRO). MVTec's single-object, fine-grained defects benefit more from LTC. Conversely, RIAD's complex industrial scenarios show greater improvements with DSSL's multi-modal semantic learning. This highlights LTC's strength in detail-focused learning and DSSL's capability in learning discriminative features from multi-modal inputs.

\newcommand{\treshl}[2]{
	{#1} \fontsize{7.5pt}{1em}\selectfont\color{mygreen}{$\!\uparrow\!$ \textbf{#2}}
}
\newcommand{\tdownreshl}[2]{
	{#1} \fontsize{7.5pt}{1em}\selectfont\color{midblue}{$\!\downarrow\!$ \textbf{#2}}
}
\newcommand{\reshl}[2]{
	\textbf{#1} \fontsize{7.5pt}{1em}\selectfont\color{mygreen}{$\!\uparrow\!$ \textbf{#2}}
}
\newcommand{\downreshl}[2]{
	\textbf{#1} \fontsize{7.5pt}{1em}\selectfont\color{midblue}{$\!\downarrow\!$ \textbf{#2}}
}
\newcommand{\greencheck}{\textcolor{green}{\ding{51}}}
\newcommand{\redcross}{\textcolor{red}{\ding{55}}}
\begin{table}[!t]
	\centering
	\small
	\renewcommand{\arraystretch}{1.}
	\renewcommand{\tabcolsep}{5.5 pt}
	\caption{\textbf{Component-wise experimental results.} }
	\begin{tabular}{cc||ccc|ccc}
		\hlineB{2.5} 
		\multirow{2}{*}{LTC} & \multirow{2}{*}{DSSL} & \multicolumn{3}{c|}{\textbf{MVTec}} & \multicolumn{3}{c}{\textbf{RIAD}}  \\ 
		\cline{3-8}
		&  & Img & Pixel &PRO& Img & Pixel &PRO \\
		\hline
		\redcross & \redcross & $90.5$ & $91.3$ & $87.3$ & $91.3$ & $93.4$ &$89.5$\\
		\redcross & \greencheck & $93.6$ & $94.3$ & $89.1$ & $92.1$ & $94.5$ & $90.8$ \\
		\greencheck & \redcross & $92.9$ & $93.7$ & $90.5$ & $93.2$ & $96.1$ & $91.5$ \\
		\rowcolor{mygray}
		\greencheck &  \greencheck & $\mathbf{95.8}$ & $\mathbf{96.1}$ & $\mathbf{91.2}$  & $\mathbf{94.9}$ & $\mathbf{98.9}$ & $\mathbf{92.3}$\\
		\hlineB{2.5} 
	\end{tabular}
	\label{tab_ablation}
\end{table}

\begin{figure}[!t]
	\centering
	\includegraphics[width=0.9\linewidth]{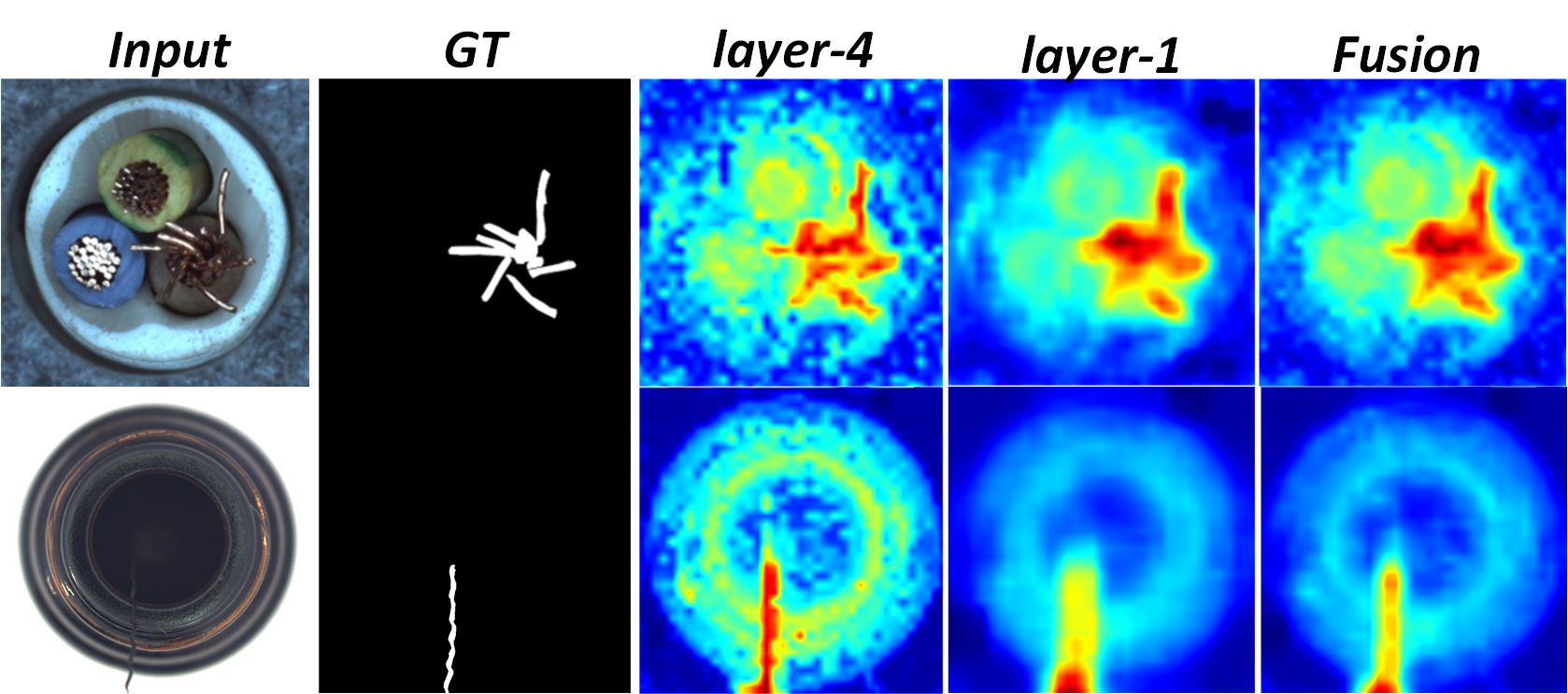}
	\caption{\textbf{LTC Multi-Level Feature Integration: attention maps from layer-1 (last) and layer-4 (fourth-to-last), and their fused result. See Sec. \ref{ablation_LTC} for ablation study}}
	\label{fig:LTC_ablation}
\end{figure}

\noindent\textbf{\emph{Ablation Study on Visual Projector:}}\label{ablation_LTC}
Tab. \ref{tab:ablation_LTC} presents the LTC's ablation results on MVTec and comparisons with various visual projectors. We set the baseline with 2$\times$ downsampled feature maps for MLP projection. Adding (\textbf{\textcolor{darkgreen}{+}}) the injection module obtains \textbf{+0.6}\%, \textbf{+1.1}\% and \textbf{+1.8}\% gains over the baseline method, respectively. Fig. \ref{fig:LTC_ablation} demonstrates the effectiveness of multi-level feature fusion. The fused map integrates layer-4's fine details with layer-1's broader structures, enhancing anomaly localization. We experiment with various projectors and keep the same settings for a fair comparison. Compared to MLP, other projectors reduce token count and significantly improve speed. Our approach surpasses the previous best method LDP-v2~\cite{chu2024mobilevlm} by \textbf{+1.4}\%, \textbf{1.8\%} and \textbf{1.3\%}, demonstrating the effectiveness of LTC. 

\begin{figure}[t]
	\centering
	\includegraphics[width=0.9\linewidth]{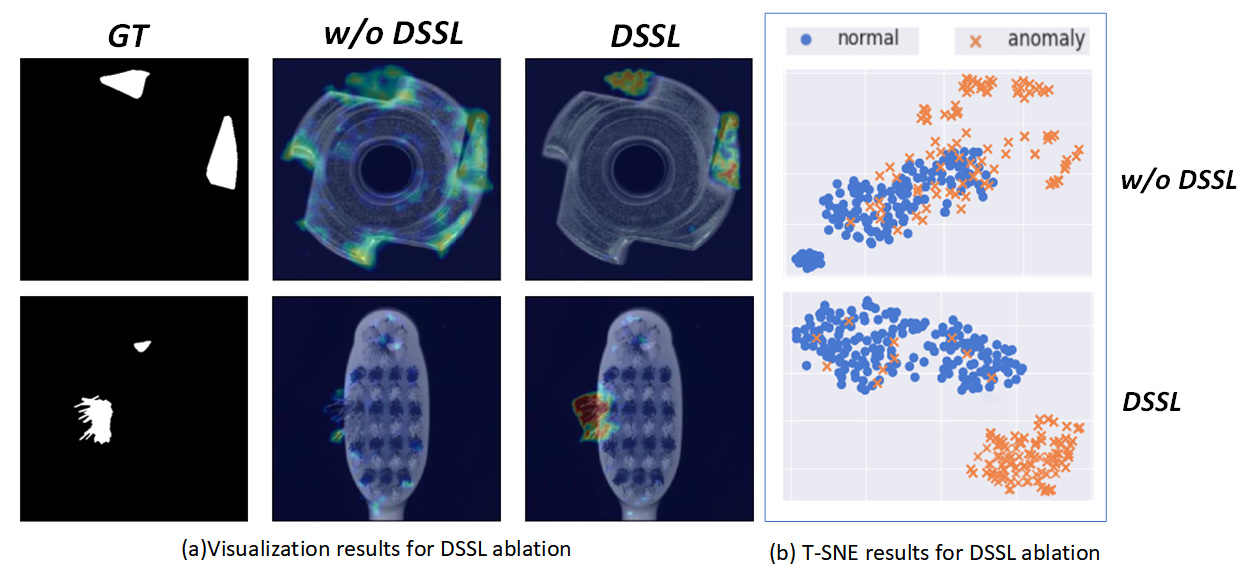}
 \vspace{-5pt}
\caption{\textbf{Ablation study on Defect-Sensitive Structure Learning module.} (Left) Visualization with and without DSSL. (Right) t-SNE plot of patch tokens. `w/o' denotes without.}
	\label{fig:DSSL_ablation}
\end{figure}

\noindent\textbf{\emph{Ablation Study on DSSL scheme.}}
Fig. \ref{fig:DSSL_ablation} presents the ablation study of Defect-Sensitive Structure Learning module. Left: Visualization demonstrates DSSL's enhanced sensitivity to anomalous structures. Right: t-SNE plot of normalized features $\Psi_{I \rightarrow T}(z)$ in LLM space (Eq. \ref{semantic_tokens}) shows improved discrimination between normal and anomalous samples through DSSL's patch-based similarity learning.
\begin{table}[t]
	\centering
	\setlength{\abovecaptionskip}{0cm}
	\captionsetup{width=0.50\textwidth}    
	\caption{\textbf{Ablation Study on Different Visual Projectors on MVTec dataset. We adopt token per second (TPS) to evaluate the throughput of LLM during inference.}}
	\label{tab:ablation-component}
	\resizebox{0.48\textwidth}{!}{
		\setlength\tabcolsep{4pt}
		\renewcommand\arraystretch{1.20}
		\begin{tabular}{l|cc|ccc}
			\hlineB{2.5}
			Method &  \#Tokens & \#TPS & {\bf Img} & {\bf Pixel}  & {\bf PRO}  \\
			\hlineB{2.5}
			\rowcolor{mygray}
			Baseline & 144 & -- & $92.9$ & $94.3$ & $89.1$  \\   
			\rowcolor{mygray}
			\textbf{\textcolor{darkgreen}{+}} {\em Injection} & 144  & -- & 93.5 & 95.4 & 90.9 \\
			\rowcolor{mygray}
			\textbf{\textcolor{darkgreen}{+}} {\em Multi-level Feature} & 144 & --& 93.8 & 95.9 & 91.0 \\
			\rowcolor{mygray}
			{\em LTC(Ours) } & 114 &  26.8  & 95.8 & 96.1 & 91.2 \\
			
			\textbf{\textcolor{darkblue}{c}} {\em MLP} & \textbf{576}  & 5.5 & \tdownreshl{91.4}{1.5} & \tdownreshl{92.6}{1.7} &  \tdownreshl{87.2}{1.9}\\
			\textbf{\textcolor{darkblue}{c}} {\em Average-Pooling }  & 144 & \textbf{31.6} & \reshl{94.9}{2.0} & \reshl{95.6}{1.3} & \reshl{91.9}{2.8} \\
			\textbf{\textcolor{darkblue}{c}} {\em Resampler} & 144 & 28.7 & \tdownreshl{91.7}{1.2} & \tdownreshl{92.4}{1.9} &  \tdownreshl{87.6}{1.5}\\
			\textbf{\textcolor{darkblue}{c}} {\em LDP-v2} & 144 & 29.8 & \tdownreshl{91.5}{1.4} & \tdownreshl{92.5}{1.8} &  \tdownreshl{87.8}{1.3}\\
			
			\hlineB{2.5} 
		\end{tabular}
	}
	\label{tab:ablation_LTC}
\end{table}

\section{Conclusion}   
In this paper, we present an MLLM-centric framework extending zero-shot anomaly detection with explainable analysis. The proposed VMAD unifies anomaly understanding, localization, and textual explanation, potentially revolutionizing industrial inspection through interpretable insights with human-AI interaction \cite{lai2024lisa,gu2024anomalygpt,nie2023reason2drive}. Besides, we collect RIAD, a comprehensive industrial dataset, supporting VQA, anomaly segmentation, and anomaly reasoning tasks, offering a versatile resource for industrial anomaly understanding. Plus, we propose novel cross-modal learning transferring patch-based visual similarity to multi-modal space for enhanced discrimination of anomalies and a novel visual projector mining multi-level features in local contexts for fine-grained defects. Comprehensive experiments on RIAD and public IAD datasets demonstrate the robustness of our method. We believe it could advance AIGC applications in automated inspection, fostering more efficient and reliable industrial processes.


\bibliographystyle{IEEEtran}
\bibliography{IEEEabrv,Refs}

\begin{thebibliography}{10}
\providecommand{\url}[1]{#1}
\csname url@samestyle\endcsname
\providecommand{\newblock}{\relax}
\providecommand{\bibinfo}[2]{#2}
\providecommand{\BIBentrySTDinterwordspacing}{\spaceskip=0pt\relax}
\providecommand{\BIBentryALTinterwordstretchfactor}{4}
\providecommand{\BIBentryALTinterwordspacing}{\spaceskip=\fontdimen2\font plus
\BIBentryALTinterwordstretchfactor\fontdimen3\font minus
  \fontdimen4\font\relax}
\providecommand{\BIBforeignlanguage}[2]{{%
\expandafter\ifx\csname l@#1\endcsname\relax
\typeout{** WARNING: IEEEtran.bst: No hyphenation pattern has been}%
\typeout{** loaded for the language `#1'. Using the pattern for}%
\typeout{** the default language instead.}%
\else
\language=\csname l@#1\endcsname
\fi
#2}}
\providecommand{\BIBdecl}{\relax}
\BIBdecl

\bibitem{yu2024tf}
Q.~Yu, K.~Zhu, Y.~Cao, F.~Xia, and Y.~Kang, ``Tf 2: Few-shot text-free
  training-free defect image generation for industrial anomaly inspection,''
  \emph{IEEE Transactions on Circuits and Systems for Video Technology}, 2024.

\bibitem{liznerski2020explainable}
P.~Liznerski, L.~Ruff, R.~A. Vandermeulen, B.~J. Franks, M.~Kloft, and K.-R.
  M{\"u}ller, ``Explainable deep one-class classification,'' \emph{arXiv
  preprint arXiv:2007.01760}, 2020.

\bibitem{li2021cutpaste}
C.-L. Li, K.~Sohn, J.~Yoon, and T.~Pfister, ``Cutpaste: Self-supervised
  learning for anomaly detection and localization,'' in \emph{Proceedings of
  the IEEE/CVF Conference on Computer Vision and Pattern Recognition}, 2021,
  pp. 9664--9674.

\bibitem{patchcore}
K.~Roth, L.~Pemula, J.~Zepeda, B.~Sch{\"o}lkopf, T.~Brox, and P.~Gehler,
  ``Towards total recall in industrial anomaly detection,'' in
  \emph{Proceedings of the IEEE/CVF Conference on Computer Vision and Pattern
  Recognition}, 2022, pp. 14\,318--14\,328.

\bibitem{radford2021CLIP}
A.~Radford, J.~W. Kim, C.~Hallacy, A.~Ramesh, G.~Goh, S.~Agarwal, G.~Sastry,
  A.~Askell, P.~Mishkin, J.~Clark \emph{et~al.}, ``Learning transferable visual
  models from natural language supervision,'' \emph{arXiv preprint
  arXiv:2103.00020}, 2021.

\bibitem{jeong2023winclip}
J.~Jeong, Y.~Zou, T.~Kim, D.~Zhang, A.~Ravichandran, and O.~Dabeer, ``Winclip:
  Zero-/few-shot anomaly classification and segmentation,'' in
  \emph{Proceedings of the IEEE/CVF Conference on Computer Vision and Pattern
  Recognition}, 2023, pp. 19\,606--19\,616.

\bibitem{zhou2023anomalyclip}
Q.~Zhou, G.~Pang, Y.~Tian, S.~He, and J.~Chen, ``Anomalyclip: Object-agnostic
  prompt learning for zero-shot anomaly detection,'' \emph{arXiv preprint
  arXiv:2310.18961}, 2023.

\bibitem{li2024clipsam}
S.~Li, J.~Cao, P.~Ye, Y.~Ding, C.~Tu, and T.~Chen, ``Clipsam: Clip and sam
  collaboration for zero-shot anomaly segmentation,'' \emph{arXiv preprint
  arXiv:2401.12665}, 2024.

\bibitem{zhu2023minigpt}
D.~Zhu, J.~Chen, X.~Shen, X.~Li, and M.~Elhoseiny, ``Minigpt-4: Enhancing
  vision-language understanding with advanced large language models,''
  \emph{arXiv preprint arXiv:2304.10592}, 2023.

\bibitem{zhang2023llava}
H.~Zhang, H.~Li, F.~Li, T.~Ren, X.~Zou, S.~Liu, S.~Huang, J.~Gao, L.~Zhang,
  C.~Li \emph{et~al.}, ``Llava-grounding: Grounded visual chat with large
  multimodal models,'' \emph{arXiv preprint arXiv:2312.02949}, 2023.

\bibitem{zhou2022MetaNet}
K.~Zhou, J.~Yang, C.~C. Loy, and Z.~Liu, ``Conditional prompt learning for
  vision-language models,'' in \emph{Proceedings of the IEEE/CVF conference on
  computer vision and pattern recognition}, 2022, pp. 16\,816--16\,825.

\bibitem{xuan2024decoupled}
S.~Xuan and S.~Zhang, ``Decoupled contrastive learning for long-tailed
  recognition,'' in \emph{Proceedings of the AAAI Conference on Artificial
  Intelligence}, vol.~38, no.~6, 2024, pp. 6396--6403.

\bibitem{meng2024deepstack}
L.~Meng, J.~Yang, R.~Tian, X.~Dai, Z.~Wu, J.~Gao, and Y.-G. Jiang, ``Deepstack:
  Deeply stacking visual tokens is surprisingly simple and effective for
  lmms,'' \emph{arXiv preprint arXiv:2406.04334}, 2024.

\bibitem{chen2024far}
Z.~Chen, W.~Wang, H.~Tian, S.~Ye, Z.~Gao, E.~Cui, W.~Tong, K.~Hu, J.~Luo, Z.~Ma
  \emph{et~al.}, ``How far are we to gpt-4v? closing the gap to commercial
  multimodal models with open-source suites,'' \emph{arXiv preprint
  arXiv:2404.16821}, 2024.

\bibitem{li2024tokenpacker}
W.~Li, Y.~Yuan, J.~Liu, D.~Tang, S.~Wang, J.~Zhu, and L.~Zhang, ``Tokenpacker:
  Efficient visual projector for multimodal llm,'' \emph{arXiv preprint
  arXiv:2407.02392}, 2024.

\bibitem{surfaceAD_survey}
V.~Zavrtanik, M.~Kristan, and D.~Sko{\v{c}}aj, ``Draem-a discriminatively
  trained reconstruction embedding for surface anomaly detection,'' in
  \emph{Proceedings of the IEEE/CVF International Conference on Computer
  Vision}, 2021, pp. 8330--8339.

\bibitem{reconstruction}
------, ``Reconstruction by inpainting for visual anomaly detection,''
  \emph{Pattern Recognition}, vol. 112, p. 107706, 2021.

\bibitem{jiang2022masked}
J.~Jiang, J.~Zhu, M.~Bilal, Y.~Cui, N.~Kumar, R.~Dou, F.~Su, and X.~Xu,
  ``Masked swin transformer unet for industrial anomaly detection,'' \emph{IEEE
  Transactions on Industrial Informatics}, vol.~19, no.~2, pp. 2200--2209,
  2022.

\bibitem{deng2024PLMNet}
H.~Deng, H.~Luo, W.~Zhai, Y.~Guo, Y.~Cao, and Y.~Kang, ``Prioritized local
  matching network for cross-category few-shot anomaly detection,'' \emph{IEEE
  Transactions on Artificial Intelligence}, vol.~5, no.~9, pp. 4550--4561,
  2024.

\bibitem{chen2024unified}
Q.~Chen, H.~Luo, C.~Lv, and Z.~Zhang, ``A unified anomaly synthesis strategy
  with gradient ascent for industrial anomaly detection and localization,''
  \emph{arXiv preprint arXiv:2407.09359}, 2024.

\bibitem{Graphcore}
G.~Xie, J.~Wang, J.~Liu, F.~Zheng, and Y.~Jin, ``Pushing the limits of fewshot
  anomaly detection in industry vision: Graphcore,'' \emph{arXiv preprint
  arXiv:2301.12082}, 2023.

\bibitem{hinton2015distilling}
G.~Hinton, O.~Vinyals, and J.~Dean, ``Distilling the knowledge in a neural
  network,'' \emph{arXiv preprint arXiv:1503.02531}, 2015.

\bibitem{zhai2023exploring}
W.~Zhai, Y.~Cao, J.~Zhang, H.~Xie, D.~Tao, and Z.-J. Zha, ``On exploring
  multiplicity of primitives and attributes for texture recognition in the
  wild,'' \emph{IEEE Transactions on Pattern Analysis and Machine
  Intelligence}, 2023.

\bibitem{deng2023PTMNet}
H.~Deng, Y.~Guo, Z.~Xu, and Y.~Kang, ``Ptmnet: Pixel-text matching network for
  zero-shot anomaly detection,'' in \emph{2023 9th International Conference on
  Big Data and Information Analytics (BigDIA)}, 2023, pp. 781--787.

\bibitem{deng2023anovl}
H.~Deng, Z.~Zhang, J.~Bao, and X.~Li, ``Anovl: Adapting vision-language models
  for unified zero-shot anomaly localization,'' \emph{arXiv preprint
  arXiv:2308.15939}, 2023.

\bibitem{cao2023segment}
Y.~Cao, X.~Xu, C.~Sun, Y.~Cheng, Z.~Du, L.~Gao, and W.~Shen, ``Segment any
  anomaly without training via hybrid prompt regularization,'' \emph{arXiv
  preprint arXiv:2305.10724}, 2023.

\bibitem{Kirillov_2023_ICCV}
A.~Kirillov, E.~Mintun, N.~Ravi, H.~Mao, C.~Rolland, L.~Gustafson, T.~Xiao,
  S.~Whitehead, A.~C. Berg, W.-Y. Lo, P.~Dollar, and R.~Girshick, ``Segment
  anything,'' in \emph{Proceedings of the IEEE/CVF International Conference on
  Computer Vision (ICCV)}, October 2023, pp. 4015--4026.

\bibitem{yin2023survey}
S.~Yin, C.~Fu, S.~Zhao, K.~Li, X.~Sun, T.~Xu, and E.~Chen, ``A survey on
  multimodal large language models,'' \emph{arXiv preprint arXiv:2306.13549},
  2023.

\bibitem{alayrac2022flamingo}
J.-B. Alayrac, J.~Donahue, P.~Luc, A.~Miech, I.~Barr, Y.~Hasson, K.~Lenc,
  A.~Mensch, K.~Millican, M.~Reynolds \emph{et~al.}, ``Flamingo: a visual
  language model for few-shot learning,'' \emph{Advances in neural information
  processing systems}, vol.~35, pp. 23\,716--23\,736, 2022.

\bibitem{li2023blip}
J.~Li, D.~Li, S.~Savarese, and S.~Hoi, ``Blip-2: Bootstrapping language-image
  pre-training with frozen image encoders and large language models,'' in
  \emph{International conference on machine learning}.\hskip 1em plus 0.5em
  minus 0.4em\relax PMLR, 2023, pp. 19\,730--19\,742.

\bibitem{liu2024visual}
H.~Liu, C.~Li, Q.~Wu, and Y.~J. Lee, ``Visual instruction tuning,''
  \emph{Advances in neural information processing systems}, vol.~36, 2024.

\bibitem{you2024ferret}
K.~You, H.~Zhang, E.~Schoop, F.~Weers, A.~Swearngin, J.~Nichols, Y.~Yang, and
  Z.~Gan, ``Ferret-ui: Grounded mobile ui understanding with multimodal llms,''
  \emph{arXiv preprint arXiv:2404.05719}, 2024.

\bibitem{lai2024lisa}
X.~Lai, Z.~Tian, Y.~Chen, Y.~Li, Y.~Yuan, S.~Liu, and J.~Jia, ``Lisa: Reasoning
  segmentation via large language model,'' in \emph{Proceedings of the IEEE/CVF
  Conference on Computer Vision and Pattern Recognition}, 2024, pp. 9579--9589.

\bibitem{rasheed2024glamm}
H.~Rasheed, M.~Maaz, S.~Shaji, A.~Shaker, S.~Khan, H.~Cholakkal, R.~M. Anwer,
  E.~Xing, M.-H. Yang, and F.~S. Khan, ``Glamm: Pixel grounding large
  multimodal model,'' in \emph{Proceedings of the IEEE/CVF Conference on
  Computer Vision and Pattern Recognition}, 2024, pp. 13\,009--13\,018.

\bibitem{gu2024anomalygpt}
Z.~Gu, B.~Zhu, G.~Zhu, Y.~Chen, M.~Tang, and J.~Wang, ``Anomalygpt: Detecting
  industrial anomalies using large vision-language models,'' in
  \emph{Proceedings of the AAAI Conference on Artificial Intelligence},
  vol.~38, no.~3, 2024, pp. 1932--1940.

\bibitem{li2023myriad}
Y.~Li, H.~Wang, S.~Yuan, M.~Liu, D.~Zhao, Y.~Guo, C.~Xu, G.~Shi, and W.~Zuo,
  ``Myriad: Large multimodal model by applying vision experts for industrial
  anomaly detection,'' \emph{arXiv preprint arXiv:2310.19070}, 2023.

\bibitem{yu2024QPN}
Y.-Q. Yu, M.~Liao, J.~Wu, Y.~Liao, X.~Zheng, and W.~Zeng, ``Texthawk: Exploring
  efficient fine-grained perception of multimodal large language models,''
  \emph{arXiv preprint arXiv:2404.09204}, 2024.

\bibitem{dong2024internlm}
X.~Dong, P.~Zhang, Y.~Zang, Y.~Cao, B.~Wang, L.~Ouyang, S.~Zhang, H.~Duan,
  W.~Zhang, Y.~Li \emph{et~al.}, ``Internlm-xcomposer2-4khd: A pioneering large
  vision-language model handling resolutions from 336 pixels to 4k hd,''
  \emph{arXiv preprint arXiv:2404.06512}, 2024.

\bibitem{radford2021learning}
A.~Radford, J.~W. Kim, C.~Hallacy, A.~Ramesh, G.~Goh, S.~Agarwal, G.~Sastry,
  A.~Askell, P.~Mishkin, J.~Clark \emph{et~al.}, ``Learning transferable visual
  models from natural language supervision,'' in \emph{International conference
  on machine learning}.\hskip 1em plus 0.5em minus 0.4em\relax PMLR, 2021, pp.
  8748--8763.

\bibitem{yu2023incremental}
Q.~Yu, K.~Zhu, W.~Wang, Y.~Cao, and Y.~Kang, ``Incremental object detection
  with image-level labels,'' \emph{IEEE Transactions on Artificial
  Intelligence}, 2023.

\bibitem{resnet}
K.~He, X.~Zhang, S.~Ren, and J.~Sun, ``Deep residual learning for image
  recognition,'' in \emph{Proceedings of the IEEE conference on computer vision
  and pattern recognition}, 2016, pp. 770--778.

\bibitem{BGAD}
\BIBentryALTinterwordspacing
X.~Yao, R.~Li, J.~Zhang, J.~Sun, and C.~Zhang, ``Explicit boundary guided
  semi-push-pull contrastive learning for supervised anomaly detection,'' 2023.
  [Online]. Available: \url{https://arxiv.org/abs/2207.01463}
\BIBentrySTDinterwordspacing

\bibitem{chen2023zero}
X.~Chen, Y.~Han, and J.~Zhang, ``A zero-/fewshot anomaly classification and
  segmentation method for cvpr 2023 vand workshop challenge tracks 1\&2: 1st
  place on zero-shot ad and 4th place on few-shot ad,'' \emph{arXiv preprint
  arXiv:2305.17382}, vol.~2, no.~4, 2023.

\bibitem{bao2023miad}
T.~Bao, J.~Chen, W.~Li, X.~Wang, J.~Fei, L.~Wu, R.~Zhao, and Y.~Zheng, ``Miad:
  A maintenance inspection dataset for unsupervised anomaly detection,'' in
  \emph{Proceedings of the IEEE/CVF International Conference on Computer
  Vision}, 2023, pp. 993--1002.

\bibitem{wang2024realiad}
C.~Wang, W.~Zhu, B.-B. Gao, Z.~Gan, J.~Zhang, Z.~Gu, S.~Qian, M.~Chen, and
  L.~Ma, ``Real-iad: A real-world multi-view dataset for benchmarking versatile
  industrial anomaly detection,'' in \emph{Proceedings of the IEEE/CVF
  Conference on Computer Vision and Pattern Recognition}, 2024, pp.
  22\,883--22\,892.

\bibitem{bai2023vision}
H.~Bai, S.~Mou, T.~Likhomanenko, R.~G. Cinbis, O.~Tuzel, P.~Huang, J.~Shan,
  J.~Shi, and M.~Cao, ``Vision datasets: A benchmark for vision-based
  industrial inspection,'' \emph{arXiv preprint arXiv:2306.07890}, 2023.

\bibitem{zhou2024pad}
Q.~Zhou, W.~Li, L.~Jiang, G.~Wang, G.~Zhou, S.~Zhang, and H.~Zhao, ``Pad: A
  dataset and benchmark for pose-agnostic anomaly detection,'' \emph{Advances
  in Neural Information Processing Systems}, vol.~36, 2024.

\bibitem{bergmann2021mvtec3d}
P.~Bergmann, X.~Jin, D.~Sattlegger, and C.~Steger, ``The mvtec 3d-ad dataset
  for unsupervised 3d anomaly detection and localization,'' \emph{arXiv
  preprint arXiv:2112.09045}, 2021.

\bibitem{zhang2022iphonescreen}
J.~Zhang, R.~Ding, M.~Ban, and T.~Guo, ``Fdsnet: An accurate real-time surface
  defect segmentation network,'' in \emph{ICASSP 2022-2022 IEEE International
  Conference on Acoustics, Speech and Signal Processing (ICASSP)}.\hskip 1em
  plus 0.5em minus 0.4em\relax IEEE, 2022, pp. 3803--3807.

\bibitem{mvtec}
P.~Bergmann, M.~Fauser, D.~Sattlegger, and C.~Steger, ``Mvtec ad--a
  comprehensive real-world dataset for unsupervised anomaly detection,'' in
  \emph{Proceedings of the IEEE/CVF conference on computer vision and pattern
  recognition}, 2019, pp. 9592--9600.

\bibitem{zou2022spot}
Y.~Zou, J.~Jeong, L.~Pemula, D.~Zhang, and O.~Dabeer, ``Spot-the-difference
  self-supervised pre-training for anomaly detection and segmentation,'' in
  \emph{European Conference on Computer Vision}.\hskip 1em plus 0.5em minus
  0.4em\relax Springer, 2022, pp. 392--408.

\bibitem{hu2021lora}
E.~J. Hu, Y.~Shen, P.~Wallis, Z.~Allen-Zhu, Y.~Li, S.~Wang, L.~Wang, and
  W.~Chen, ``Lora: Low-rank adaptation of large language models,'' \emph{arXiv
  preprint arXiv:2106.09685}, 2021.

\bibitem{chu2024mobilevlm}
X.~Chu, L.~Qiao, X.~Zhang, S.~Xu, F.~Wei, Y.~Yang, X.~Sun, Y.~Hu, X.~Lin,
  B.~Zhang \emph{et~al.}, ``Mobilevlm v2: Faster and stronger baseline for
  vision language model,'' \emph{arXiv preprint arXiv:2402.03766}, 2024.

\bibitem{nie2023reason2drive}
M.~Nie, R.~Peng, C.~Wang, X.~Cai, J.~Han, H.~Xu, and L.~Zhang, ``Reason2drive:
  Towards interpretable and chain-based reasoning for autonomous driving,''
  \emph{arXiv preprint arXiv:2312.03661}, 2023.

\end{thebibliography}

\end{document}